\documentclass{article} 
\usepackage{collas2023_conference,times}
\usepackage{easyReview}

\usepackage{hyperref}
\usepackage{url}
\usepackage{graphicx}
\usepackage{wrapfig}
\usepackage{xcolor}
\usepackage{subcaption}
\usepackage{placeins}
\usepackage{stmaryrd}
\usepackage{booktabs}
\usepackage{pgfplots}
\usepackage{amsthm}
\usepackage{algorithm}
\usepackage{algorithmicx}
\usepackage{algpseudocode}
\usepackage{layouts}
\usepackage{graphics}
\usepackage{tikz}
\usepackage{wrapfig}
\usetikzlibrary{matrix}

\usepackage{amsmath,amsfonts,bm}

\def\eqref#1{equation~\ref{#1}}

\def\1{\bm{1}}

\DeclareMathAlphabet{\mathsfit}{\encodingdefault}{\sfdefault}{m}{sl}
\SetMathAlphabet{\mathsfit}{bold}{\encodingdefault}{\sfdefault}{bx}{n}

\def\sX{{\mathbb{X}}}

\newcommand{\E}{\mathbb{E}}

\newcommand{\R}{\mathbb{R}}

\newcommand{\Var}{\mathrm{Var}}

\DeclareMathOperator*{\argmax}{arg\,max}

\usepackage{hyperref}
\hypersetup{
    colorlinks=true,
    linkcolor=red,
    filecolor=magenta,
    urlcolor=blue,
    citecolor=purple,
    pdftitle={Overleaf Example},
    pdfpagemode=FullScreen,
    }

\title{VIBR: Learning View-Invariant Value Functions for Robust Visual Control}

\author{Tom Dupuis$^{1,2}$, Jaonary Rabarisoa$^{1}$, Quoc-Cuong Pham$^{1}$, David Filliat$^{2,3}$ \\
$^1$Universite Paris-Saclay, CEA, List, F-91120, Palaiseau, France ´\\
$^2$U2IS, ENSTA Paris, Institut Polytechnique de Paris, Palaiseau, France \\
$^3$INRIA FLOWERS \\
\texttt{tom.dupuis@cea.fr} 
}

\newtheorem{prop}{Proposition}[section]

\theoremstyle{definition}
\newtheorem{definition}{Definition}[section]
\newtheorem{assumption}{Assumption}[section]

\collasfinalcopy 

\begin{document}

\maketitle

\begin{abstract}
End-to-end reinforcement learning on images showed significant progress in the recent years. Data-based approach leverage data augmentation and domain randomization while representation learning methods use auxiliary losses to learn task-relevant features. Yet, reinforcement still struggles in visually diverse environments full of distractions and spurious noise. In this work, we tackle the problem of robust visual control at its core and present VIBR (View-Invariant Bellman Residuals), a method that combines multi-view training and invariant prediction to reduce out-of-distribution (OOD) generalization gap for RL based visuomotor control. Our model-free approach improve baselines performances without the need of additional representation learning objectives and with limited additional computational cost. We show that VIBR outperforms existing methods on complex visuo-motor control environment with high visual perturbation. 
Our approach achieves state-of the-art results on the Distracting Control Suite benchmark, a challenging benchmark still not solved by current methods, where we evaluate the robustness to a number of visual perturbators, as well as OOD generalization and extrapolation capabilities.
\end{abstract}

\section{Introduction}
\label{sec:introduction}

Learning policies that are invariant to visual distractions is crucial for the usage of reinforcement learning 
for real-world visuomotor control problems. The visual variance of the real world is practically unbounded and
the distribution of possible events has an extremely heavy tail. For example, end-to-end autonomous driving struggles with 
the never-ending list of edge cases sparsely present in the data. Although there is real progress in visual 
generalization in robotic manipulation for example, we are quite far from reaching human levels of robustness.\

Data augmentation is extensively used for building inductive biases in pure computer vision tasks, such as image 
classification. One can not imagine reaching state-of-the-art performance on usual benchmarks without using a careful combination 
of transformation on images. In particular, pretraining representations with teacher-student self-supervised objectives
is a popular and successful method. Such architecture usually enforces invariance of representations
to different augmentations of the same image. As such, multiple works take inspiration from success in computer vision and 
build methods to learn visual invariances for control and reinforcement learning~\citep{yarats2020image}.\

In the case of classification, invariance of representations is a reasonably efficient optimization strategy, as 
classifying images from pretrained representations is relatively straightforward and only require linear probing 
in most cases. Because classifying the content of an image is a semantically high level task, the class label is resilient 
to a lot of intense visual transformation of the image. Features such as exact position, relative organization 
and textures of entities in the image are usually not predictive of the class label. Data augmentation for
implicit invariance is straightforward to apply in these cases and computer vision pipelines fully take advantage
of this fact.

In the case of control, however, finding a meaningful and useful policy 
from a given state representation is multiple orders of magnitude more complex than assigning a class token. 
Small visual changes in images might necessitate very different action decisions, which means policies must be 
sensitive to high-frequency variations in images. This goes against principles of data augmentation in computer 
vision tasks like classification where invariances holds for very aggressive augmentations. The problem gets worse when
policies are implemented with deep neural networks, which are known to learn over low-frequency components of the
data~\citep{rahaman2019spectral}. This directly suggests that any representational approximation error will directly impact the upper-bound performance
of the policy. The problem is slightly mitigated with discrete action which theoretically allow partial mode collapse 
of observations in the latent space, but continuous control represents a serious challenge.

\begin{figure}
    \begin{center}
        \includegraphics[width=.7\textwidth]{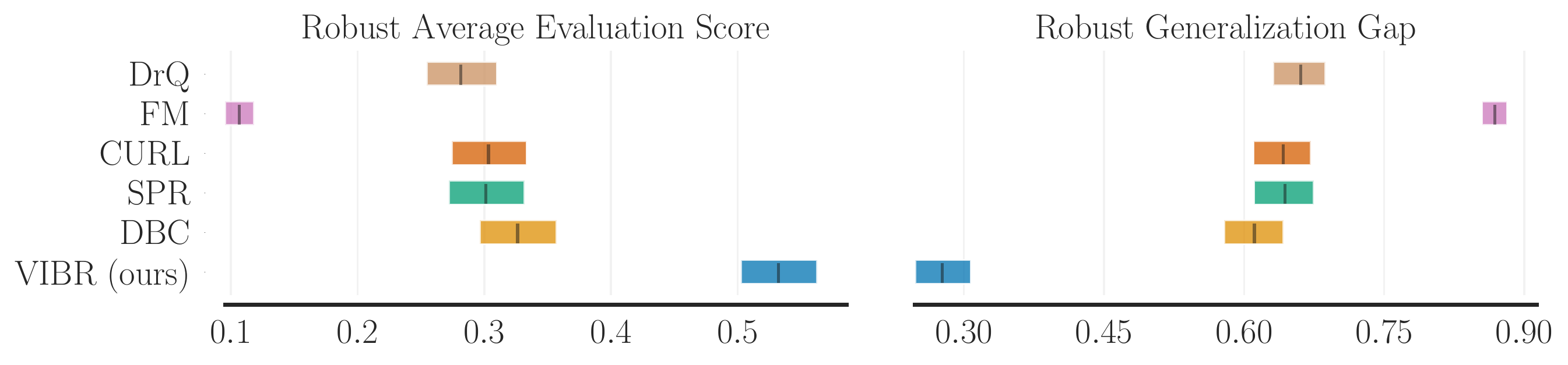}
    \end{center}
    \caption{Evaluation metrics at the end of training aggregated over all 5 curriculum benchmarks and 6 tasks of the Distracting Control Suite, 21 episodes and 4 seeds each. See section~\ref{sec:results} for details.}
    \label{fig:iqm}
\end{figure}

From this observation, we present \textbf{View-Invariant Bellman Residuals} (\textbf{VIBR}) for robust visuomotor 
control under visual distractions. Importantly, our approach does not require a representation learning self-supervised 
loss and directly performs invariant prediction, which we show is a relaxed constraint that enables better optimization 
and performances.
We show that VIBR is able to efficiently train an agent with visual generalization 
capabilities without losing on convergence speed and asymptotic performance on the original task. In particular, we 
derive a view-invariant temporal-difference loss combining multi-view empiricial risk minimization with variance 
regularization. We show that view-invariant prediction can serve as a powerful inductive bias for learning robust policies
and value functions with reinforcement learning, even under intense perturbations. Our main contributions are:

\begin{itemize}
    \item a novel methodology for robust visuomotor control based on temporal-difference learning on images using 
    invariant prediction principles
    \item empirical results on the Distracting Control Suite benchmark \citep{stone2021distracting} with 
    state-of-the-art results on raw training performance (Figure~\ref{fig:iqm}) and out-of-distribution (OOD) generalization under the 
    hardest setting of dynamic distractions
    \item a fair comparison with competitive baselines of invariant representation learning for RL 
    \item an analysis of the influence of inter-view variance regularization in learning dynamics
\end{itemize}

\section{Context and Problem Setting} \label{sec:preliminaries}

We first details the context and notations, then position ourselves compared to representation learning and introduce the tools of risk minimization.

\subsection{Value-Based Reinforcement Learning}

\paragraph{Markov Decision Process and RL:}
We define the Markov Decision Process $\mathcal{M}=\langle \mathcal{S},\mathcal{A},P,R,\gamma \rangle$ where $S$ 
is the set of states, $A$ the set of actions, $P: \mathcal{S} \times \mathcal{A} \rightarrow \mathcal{S}$ 
the transition probability function, $R: \mathcal{S} \times \mathcal{A} \times \mathcal{S} \rightarrow \R$ 
the reward function and $\gamma$ a discounting factor for discriminating between short-term and long-term rewards. 
We also define the transition tuple containing state, action, reward and next state
as $T=(s, a, r, s+)$ where $a \sim \pi$, $s+ \sim P^\pi(s+|s)$ is the successor state of $s$ and 
$r \sim R(s,a,s+)$ the reward. In this setting, 
reinforcement learning aims to maximize the total reward received by the agent. Mathematically, it corresponds 
to finding a policy $\pi$ that maximizes the 
(discounted) expected return  $\E_{T \sim \pi, P^\pi} [\sum_{t=0}^{\infty} \gamma^t r_t]$.\

\paragraph{Bellman operator and Bellman error:} 
Value-based algorithms for control estimate the (state-action) \textit{value function} $Q^\pi$ 
which is defined for every state $s_t$: 
$Q^\pi(s_t,a_t) = \E_\pi \left[ \sum_{k=0}^{+\infty} \gamma^k r_{k+t+1} \right]$ with $r_{k+t+1} = R(s_t,a_t,s+)$.
We will use the notation $Q(s,\pi) = \E_{a \sim \pi} \left[ Q(s,a) \right]$ for simplicity. This connects the state value 
function to the state-action value function: $V^\pi(s) = Q(s,\pi)$.
We can define the Bellman evaluation operator $\mathcal{T}^\pi$:
\begin{equation}
    \label{equ:bellman}
    \mathcal{T}^\pi Q^\pi(s,a) = r(s,a) + \gamma \E_{s+ \sim P^\pi(s+|s)}Q(s+,\pi)
\end{equation}
We define the \textbf{Bellman error} and \textbf{Bellman residuals}:
\begin{equation}
    \label{equ:bellman_error}
    (\mathcal{B}^\pi Q)(s,a) = Q^\pi(s,a) - \mathcal{T}^\pi Q^\pi(s,a) 
\end{equation}
\begin{equation}
    \label{equ:bellman_residual}
    \mathcal{L}_\mathrm{BR} = \E_{T} \left[ \lvert\lvert (\mathcal{B}^\pi Q)(s,a) \rvert\rvert^2 \right]
\end{equation}

\subsection{Observations and Representations}
\label{sec:Observers}
\paragraph{Observer:}
We assume the Block-MDP setting as defined by \cite{du2019provably}.
\begin{definition}{\textbf{BlockMDP}}
    \label{def:BlockMDP}
    A block MDP is the tuple $\mathcal{M_x} = \langle \mathcal{S},\mathcal{A},\mathcal{O},P,x,R,\gamma \rangle$ which 
    is an extension of the MDP defined above
    where $\mathcal{O}$ is an observation space (potentially much bigger than $\mathcal{S}$)
    and $x: \mathcal{S} \rightarrow \mathcal{O}$ is a "context-emission function" or \textbf{observer}.
\end{definition}

The \textit{observer} $x$ is responsible for generating observation for each state of the system but is most of the time
unknown. We equip the Block MDP with the following assumption:

\begin{assumption}{\textbf{Block structure}}
\label{ass:BS}
    Each observation $o$ uniquely determines its generating state $s$. That is, the observation space $\mathcal{O}$ 
    can be partitioned into disjoint blocks $O_s$, each containing the support of the conditional distribution $x(\cdot|s)$ \citep{du2019provably}.
\end{assumption}
This ensures that for a given BMDP $\mathcal{M}_x$, the corresponding observer $x$ is well-defined and injective and  
guarantees non-ambiguity of observations
This assumption gives us the Markov property and makes $\mathcal{M}_x$ a proper MDP on which we can work and use all the existing results of RL.\

We now have a constructive method to define multiple \textit{views} or observations of a single state. All it requires is 
to sample multiple observers but otherwise keeping other elements of the MDP constant. Given a list of observers 
$\left[x^1,...,x^K\right]$, we have a list of MDPs $\left[\mathcal{M}_{x^1},...,\mathcal{M}_{x^K}\right]$ to train on. 
Importantly, each of these MDP share the same reward function, dynamics and state and action space and thus encode the same
exact task. If the environments are run in parrallel, this provides extremely rich information to extract invariance to 
observers, as the hidden state at each step is equal over all environments. Finding optimal policies is then only a matter of information retrieval from observations and should be theoretically
possible with any view.\

\paragraph{Representation learning:}
Numerous works take a different view to robustness on visual variations and focus on guiding the parameters of the 
value network with the help of auxiliary tasks \citep{jaderbergreinforcement, bellemare2019geometric, dabney2021value}.
In this context, the approximated value function can be decomposed with a representation network $\phi$ and a 
linear layer $w$: $Q^\pi_\theta(s, a) = \phi(s,a) \cdot w$\footnote{Note that we could arbitrarily choose the representation layer 
to be earlier than the penultimate layer and the linear layer would become a shallow non-linear network, but this will not change our argument}.\

Representation learning aims to learn $\phi$ with some auxiliary objectives to better condition 
the space of value functions to accelerate learning and/or improve generalization. With online reinforcement learning, 
the agent is trained on both RL and (self-supervised) representation learning objectives at the same time: 
$\mathcal{L}_{\mathrm{tot}} = \mathcal{L}_{\mathrm{aux}}(\phi) + \mathcal{L}_{\mathrm{RL}}(\phi,w)$
where $\mathcal{L}_{\mathrm{aux}}$ is the (self-supervised) representation loss and $\mathcal{L}_{\mathrm{RL}}$ is the 
RL loss.\

\subsection{Multi-Domain Training}

We are interested in training value functions that are robust to out-of-distribution domain shifts for robust visual control. 
Given $K$ domains $\mathcal{D}^k$, we can define the empirical risk associated to each domain as the expectation of a loss
function $l$ on this domain: $\mathcal{R}(\mathcal{D}^k;\theta) = \E_{X \sim \mathcal{D}^k} [l(X;\theta)]$ where $X$ is the 
training data containing individual samples and $\theta$ is the parameters of the model being trained. A simple approach 
is to perform \textbf{Empirical Risk Minimization} (ERM), i.e. averaging risks on training domains:
\begin{equation}
    \theta^\ast \in \min \sum_k \lvert \mathcal{D}^k \rvert \mathcal{R}(\mathcal{D}^k;\theta)
\end{equation}

This approach doesn't guarantee transfer under OOD conditions. Given different domains
$\left[\mathcal{D}^1, ... \mathcal{D}^K\right]$, it is possible to minimize the ERM objective by overfitting on one particular domain $\mathcal{D}^k$
while not optimizing a lot over other domains, let alone unknown domains. To prevent that, it is possible to use constraint optimization to force equality of training risks across domains:
\begin{equation}
    \min_{\theta} \sum_k \lvert \mathcal{D}^k \rvert \mathcal{R}(\mathcal{D}^k;\theta) \quad
 \textrm{s.t.} \quad \forall (k,l) \in [1...K] \times [1...K] \quad \mathcal{R}(\mathcal{D}^k;\theta) = \mathcal{R}(\mathcal{D}^l;\theta)
\end{equation}

The constraint of perfect equality of risk is equivalent to enforcing variance of risks to zero:
\begin{equation}
    \min_{\theta} \quad \sum_k \lvert \mathcal{D}^k \rvert \mathcal{R}(\mathcal{D}^k;\theta) \quad
     \textrm{s.t.} \quad  \Var(\mathcal{R}(\mathcal{D}^k;\theta)) = 0
\end{equation}

This hard constraint can then be relaxed with a soft convex penalty facilitating optimization, leading to the V-REx approach \citep{krueger2021out}:
\begin{equation}
    \min_{\theta} \quad \sum_k \lvert \mathcal{D}^k \rvert \mathcal{R}(\mathcal{D}^k;\theta) + \beta \Var(\mathcal{R}(\mathcal{D}^k;\theta))
\end{equation}
\section{Learning View-Invariant Value Functions}

\subsection{Invariant Representation Learning and Minimal Constraints}
\label{meth:cosine}

Before introducing our VIBR approach, we will clearly formalize our objective and see how it is different from the representation learning approach.

\begin{definition}{\textbf{View-invariant functions}} 
    \label{View-invariance}
    A real-valued function $f: \mathcal{O} \rightarrow \R$ is said to be \textit{view-invariant} 
    if it is invariant to any observer transformation (or "observation") on the state space.
    Formally, for all state $s$ in $\mathcal{S}$ and observers $(x,x')$ in $\sX$ (the set of all possible observers), we have:
    $$f(x(s)) = f(x'(s))$$
\end{definition}

Thanks to assumption \ref{ass:BS}, states can be uniquely recovered from their observations and we
simplify the definition with a functional equation $f(x)=f(x')$ with no ambiguity. 
We will use this notation further for clarity of writing.\

Let's consider the family of \textit{realizable} value functions parametrized by neural networks:
\begin{equation}
    \mathcal{Q}_\Theta
    = \left\{Q_\theta^\pi: \mathcal{O} \times \mathcal{A} \rightarrow \R \quad \text{s.t.} 
    \quad \forall o \in \mathcal{O}, a \in \mathcal{A},  \quad Q_\theta^\pi(o,a)= \mathcal{T}^\pi Q_\theta^\pi(o,a) \right\}
\end{equation}
By definition, these value function have a Bellman error (Eq. \ref{equ:bellman_error}) of zero and exactly fullfil Bellman equation. 
This value functions evaluates real policies in the given MDP with no approximation error. We will relax this assumption 
further in the discussion. 

\begin{definition}{\textbf{View-invariant value functions}}
    \label{def:view-invariant}
    The set of \textit{view-invariant value functions} is the subset of realizable value functions that are view-invariant.
    \begin{equation}
        \label{equ:VI-value}
        \mathcal{Q}_\Theta^{\mathrm{inv}} 
        = \left\{Q_\theta^\pi \subset \mathcal{Q}_\Theta \quad \text{s.t.} \quad 
        \forall (x,x') \in \sX, \quad Q_\theta^{\pi,\mathcal{M}_x}= Q_\theta^{\pi,\mathcal{M}_{x'}} \right\} 
    \end{equation}
\end{definition}

This is exactly what we are looking for: such value functions would completely ignore spurious visual details introduced by
observers and only extract the true hidden state from the observation. 

However, many work engage with this problem with a more constrained approach by learning implicit
invariant \textit{representations}:

\begin{definition}{\textbf{Representation-invariant value functions}}
    This is the set of realizable value functions with view-invariant intermediary representations:
    \begin{equation}
        \mathcal{Q}_{\Phi,\mathrm{w}}^{\mathrm{inv}} = \left\{Q_\theta^\pi:=\phi \cdot w \quad \text{s.t.} 
        \quad \forall (x,x') \in \sX, \quad \phi(x)= \phi(x') \right\}
    \end{equation}
    with $\phi$ and $\mathrm{w}$ defined in section \ref{sec:Observers}.
\end{definition}

\paragraph{Minimal representation constraints:}
We immediately have the following inclusion: $\mathcal{Q}_{\Phi,\mathrm{w}}^{\mathrm{inv}} \subset \mathcal{Q}_\Theta^{\mathrm{inv}}$. 
Indeed, if $\phi(x)= \phi(x')$, then 
$\phi(x) \cdot w= \phi(x')\cdot w$, which proves the inclusion. The inverse is not true: invariance of representations
is a stricter condition on the function than invariance of value prediction on two aspects. First, if we consider the natural
assumption of neural networks with finite capacity, then invariant representations imply that $\phi$ has less parameter
available to both satisfy the constraint and provide good features for the last layer $w$ to perform value estimation. Secondly,
from an optimization perspective, estimating a scalar value is easier than a full latent vector representation. Representation 
learning is akin to model learning and world models can be more complex functions than actual optimal policies 
(and their value functions) because you need to learn the entire dynamics of the environment which might not be necessary to predict 
acurate values and learn good policies.\

\paragraph{Optimizability of representation learning:} 
Let's now remove the hypothesis of perfect approximation and suppose we have a non-zero Bellman error 
(or stricly positive Bellman residuals). By constraining the representation with an auxiliary objective, 
the network must now solve two tasks at once: producing view-invariant representation and giving accurate 
representations for value estimation. In theory, having a perfectly view-invariant representation
is enough to guarantee view-invariant value function. However, in practice models capacities are finite, 
gradients are approximated through sampling and loss are never minimized to zero. The network will operate a trade-off between RL and
representation objective if they are not aligned enough. Moreover, representation errors might compound with RL errors and put an
upper bound on the maximally achievable performance. If the learning objective is particularly noisy and hard to optimize, 
the approximation error of representation learning might become prohibitive of any progress on the RL objective. This problem might be less sensitive if we only seek view invariance instead of representation invariance. We empirically 
demonstrate this intuition in the experiment section.\

\subsection{View-Invariant Bellman Residuals}

Our goal is to solve these limitations by finding a better optimization objective to learn a view-invariant value function. 
Following our discussion, we relax the invariant representation assumption and place ourselves instead in a purely
invariant prediction setting. We wish to attain invariant prediction in an end-to-end manner, which would let the model 
learn only to use necessary and sufficient information to solve the task, without intermediate step.\


\begin{figure}[t!]
    \centering
    \begin{subfigure}[t]{0.33\textwidth}
        \centering
        \scalebox{0.5}{
            \begin{tikzpicture}
            
                \matrix [matrix of nodes,         nodes={minimum size=8mm,outer sep=0pt},         column sep=2mm,         row sep=2mm,         anchor=north west] (legend1) at (-2.5,5)
                {
                |[fill=red!20]| \\
                |[fill=yellow!20]| \\
                };
                
                \matrix [matrix of nodes,         nodes={minimum size=8mm,outer sep=0pt},         column sep=2mm,         row sep=2mm,         anchor=north west] (legend2) at ([yshift=-1cm]legend1.south west)
                {
                |[fill=orange!20]| \\
                |[fill=blue!20]| \\
                };
                
                \draw [dashed] ([shift={(-0.5cm,0.5cm)}]legend1.north west) rectangle ([shift={(0.5cm,-0.5cm)}]legend1.south east);
                
                \draw [dashed] ([shift={(-0.5cm,0.5cm)}]legend2.north west) rectangle ([shift={(0.5cm,-0.5cm)}]legend2.south east);
                
                \node [above] at (legend1.north) {Train};
                \node [above] at (legend2.north) {Test};
            
                \draw[line width=2cm,red!20] (2,0) --( 2,5);
                \draw (2, 4) node[below, black, font=\large] {$\operatorname{Span}(\mathbf{x^k})$};
                \draw[line width=2cm,yellow!20] (5,0)--(5,5);
                \draw (5, 4) node[below, black, font=\large] {$\operatorname{Span}(\mathbf{x^l})$};
                \draw[line width=1cm,orange!20] (3.5,0)--(3.5,5);
                \draw (3.5,0) node[below, orange, font=\small] {Interpolation};
                \draw[line width=1cm,blue!20] (0.5,0)--(0.5,5);
                \draw (0.5,0) node[below, blue!50, font=\small] {Extrapolation};
                \draw[line width=1cm,blue!20] (6.5,0)--(6.5,5);
                \draw (6.5,0) node[below, blue!50, font=\small] {Extrapolation};
                
                \draw[very thick,-latex] (0, 0) -- (7, 0) node[right, font=\large] {$\mathcal{O}$};
                \draw[very thick,-latex] (0, 0) -- (0, 5) node[above, font=\large]
                {$\mathcal{L}_{\mathrm{VIBR}}$};
                
                \draw [ultra thick,blue] plot [smooth, tension=0.6] coordinates { (0,4) (1,1) (2,1) (3,1) (3.5,3.5) (4,2) (5,2) (6,2) (7,5)} node[right, blue] {$\beta=0$};
                \draw [ultra thick,red,dashed] plot [smooth, tension=0.6] coordinates { (0,2.25) (1,1.5) (3,1.5) (3.5,2) (4,1.75) (6,1.75) (7,2.5)} node[right, red] {$\beta>0$};
                \draw [very thick, orange] (0,2.0) -- (7,2.0) node[right, orange] {$\beta \rightarrow +\infty$} ;
                
                \draw[very thick, black, dotted] (1,-0.1) -- (1,5);
                \draw[very thick, black, dotted] (3,-0.1) -- (3,5);
                \draw[very thick, black, dotted] (4,-0.1) -- (4,5);
                \draw[very thick, black, dotted] (6,-0.1) -- (6,5);
            
            \end{tikzpicture}
            }
        \caption{}
        \label{fig:schematics}
    \end{subfigure}
    \hfill
    \begin{subfigure}[t]{0.6\textwidth}
        \centering
        \includegraphics[width=\textwidth]{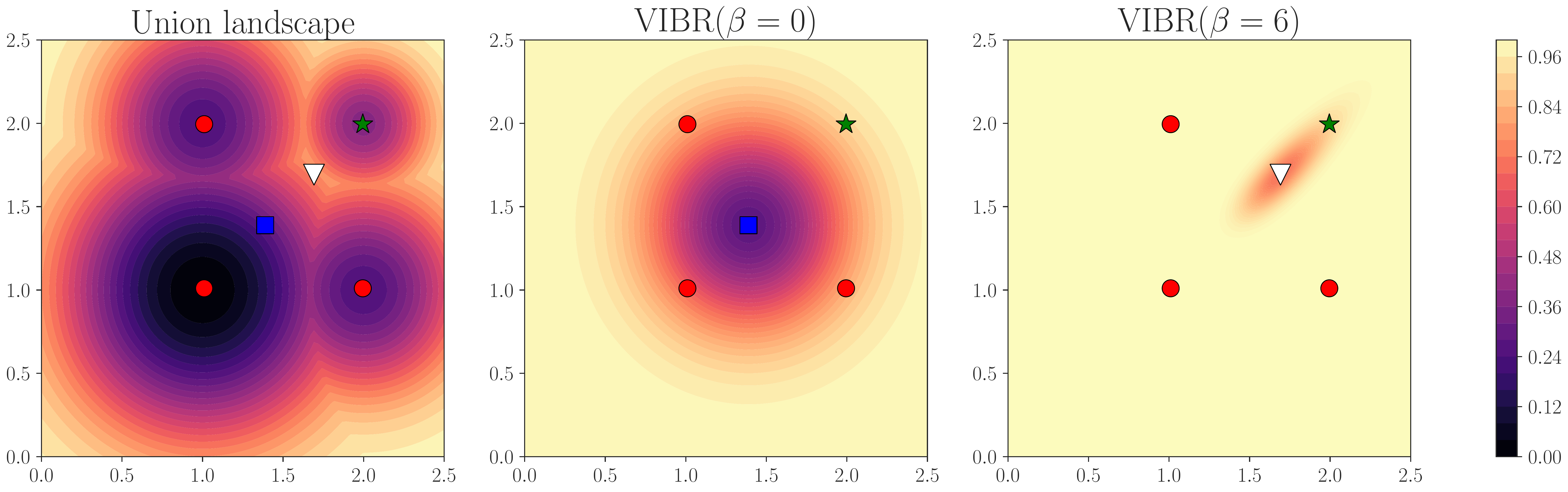}
        \caption{}
        \label{fig:loss_landscape}
    \end{subfigure}
   \caption{\textbf{(a):} \textbf{Loss landscape of VIBR in observation space}. Given two observers $x^k, x^l$ that define training domains in $\mathcal{O}$, VIBR uses V-REx to control the ID (interpolation) and OOD (extrapolation) risks \textbf{(b):} \textbf{Toy Experiment of VIBR loss landscape in parameter space} \textcolor{red}{Red points} are individual local minima of each training domains (3). \textcolor{green}{Green star} is individual minimum of the testing domain (held-out). \textcolor{blue}{Blue square} is the global minimum of ERM over training domains. \textbf{White triangle} is the global minimum of V-REx over training domains. See Appendix \ref{appendix:toyExp} and Section \ref{sec:toyexp} for details.}
  \end{figure}
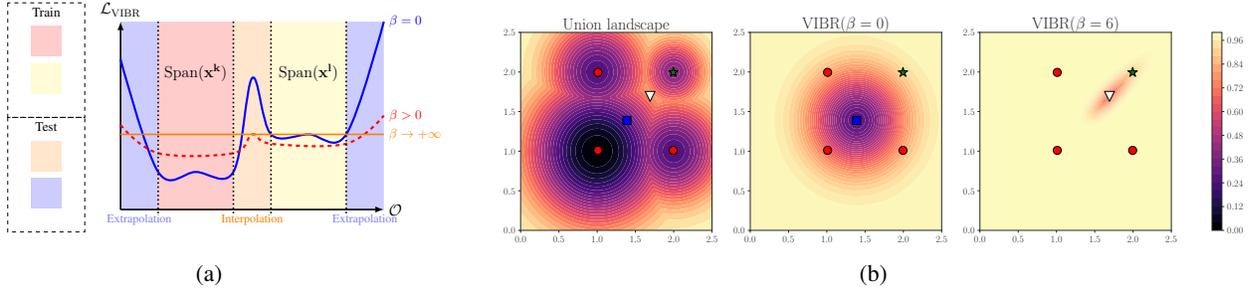

We first begin by observing the following property: 
\begin{prop}
    Suppose $Q_\theta^\pi$ a parametrized value function. $Q_\theta^\pi \in {Q}_\Theta^{\mathrm{inv}}$ if and only if: 
    $$\forall (x,x') \in \sX , s \in \mathcal{S}, a \in \mathcal{A} \qquad 
    Q_\theta^{\pi}(x(s),a) = \mathcal{T}^\pi Q_\theta^{\pi}(x'(s),a)$$
\end{prop}
The proof is immediate by noticing that for $Q_\theta^\pi \in {Q}_\Theta^{\mathrm{inv}}$, the following equalities hold:
\begin{equation*}
    \forall (x,x') \in \sX , s \in \mathcal{S}, a \in \mathcal{A} \qquad 
    Q_\theta^{\pi}(x(s),a)= Q_\theta^{\pi}(x'(s),a) = \mathcal{T}^\pi Q_\theta^{\pi}(x'(s),a)
\end{equation*}

Because this property holds regardless of the observer, this directly gives us a single unified objective that optimizes both 
for value convergence and view-invariance: 
\begin{equation}
    \theta^\ast  \label{equ:min_bellman_error}
                 = \min_\theta \E_{(x,x') \in \sX} E_{(s,a)} \left[ Q_\theta^{\pi}(x(s),a) - \mathcal{T}^\pi Q_\theta^{\pi}(x'(s),a) \right] = \min_\theta \E_{(x,x') \in \sX} \left[ \mathcal{B}^\pi Q_\theta(x,x') \right]
\end{equation}
The second equality simply defines the notation $\mathcal{B}^\pi Q_\theta(x,x')$. For better optimization, we replace the Bellman error with the Bellman residuals:
\begin{equation}
    \mathcal{L}_\mathrm{BR}(k,l) : = \lVert \mathcal{B}^\pi Q_\theta(x^k,x^l) \rVert ^2
\end{equation}
with $\left[ x^1, ... x^K\right]$ a set of sampled observers that will allow us to extract multiple views from the same scene.
We then perform empirical Bellman residuals minimization and approximate the expectation of equation~\ref{equ:min_bellman_error} with an empirical average using sampled observers:
\begin{equation}
    \widehat{\E}\left[ \mathcal{L}_\mathrm{BR}(k,l)\right] = \frac{1}{K^2}\sum_{k,l}\mathcal{L}_\mathrm{BR}(k,l)
\end{equation}

\label{meth:OOD}

In order to improve out-of-distribution generalization, we finally add a soft convex penalty with variance, following the V-REx approach. The final objective becomes:
\begin{equation}
    \begin{aligned}
        & \mathcal{L}_\mathrm{VIBR} = \widehat{\E}_{(k,l)}\left[ \mathcal{L}_\mathrm{BR}(k,l)\right] 
        + \beta \widehat{\Var}(\mathcal{L}_\mathrm{BR}(k,l)) \\
        &\textrm{where } \quad \widehat{\Var}(\mathcal{L}_\mathrm{BR}(k,l)) = \frac{1}{K^2}\sum_{k,l} \left( \mathcal{L}_\mathrm{BR}(k,l) - \widehat{\E}\left[ \mathcal{L}_\mathrm{BR}(k,l)\right] \right)^2
    \end{aligned}
\end{equation}
The detailed usage of VIBR in conjunction with Q-learning is described in algorithm \ref{alg:VIBR}.

\begin{algorithm}
    \caption{View-Invariant Bellman Residuals for Q-learning}
    \label{alg:VIBR}
    \begin{algorithmic}[1]
        \State \textbf{Initialize} Network parameters $\theta$, $K$ observers $\left[x^1, ..., x^K \right]$,
        replay buffer $\mathbb{B}$, variance reg. hyperparam $\beta$
        \For{episode $= 1, M$}
            \For{timestep $= 1, T$}
                \State Get the views from observers: $\forall k \in [1,K] \quad o_t^k=x^k(s_t)$
                \State Choose action with ensembling $a_t=\argmax_a \frac{1}{K} \sum_{k=1}^K Q^\pi_\theta(o_t^k,a)$
                \State Add transition $T_t=(\left[o_t^{1:K}\right], a_t, r_t, \left[o_{t+1}^{1:K}\right])$ to replay buffer
                \State Sample a batch of transitions $T_i \sim \mathbb{B}$ 
                \State Compute observer-pairwise Bellman residuals $$\forall (k,l) \in [1,K]^2 \quad \mathcal{L}_\mathrm{BR}(k,l)=
                        \E_{T_i} \left[ \lvert \mathcal{B}^\pi Q_\theta(T_i) \rvert^2 \right]$$
                \State Compute VIBR loss: 
                $$\mathcal{L}_\mathrm{VIBR} = \Hat{\E}\left[ \mathcal{L}_\mathrm{BR}(k,l)\right] + \beta \Hat{\Var}(\mathcal{L}_\mathrm{BR}(k,l))$$
                \State Update Q-network parameters $\theta \gets \theta - \alpha \nabla_\theta \mathcal{L}_\mathrm{VIBR}$
                \State (Optional) update target parameters $\Bar{\theta} \gets \tau \theta + (1-\tau) \Bar{\theta}$
            \EndFor
        \EndFor
    \end{algorithmic}
\end{algorithm}
\section{Experiments}
\label{sec:results}
\subsection{Variance Reduction Toy Experiment}
\label{sec:toyexp}

We empirically validate our assumptions of better generalization with a small toy experiment by visualizing the loss landscape of a 2-parameter model. We simulate training VIBR with different observers by creating four distincts local minima with 
contiguous valleys, corresponding to three different training domains (minimum at red dots) and one testing domain (minimum at green star) by analogy. 
On Figure \ref{fig:loss_landscape} left, we show the joint loss landscape of these four domains. All four domains have different minimum and local curvature to simulate asymmetry in optimization difficulty. The goal is to train the model to perform well on the unseen test domain.
We compare what the loss landscape would be with $\beta=0$ or $\beta>0$ using VIBR in Figure \ref{fig:loss_landscape}
center and right. The first case 
is equivalent to ERM over training domains. We suppose that we have equal sampling of data points for each domain, hence equal weight. This makes the global minimum of ERM attracted to the bottom/left training domain (red dot at 1,1 coordinates)
which has a big impact on the ERM loss. Intuitively, this illustrates overfitting to one particular domain in the 
case of using a simple empirical average over training observers with VIBR($\beta=0$). 
The second case corresponds with regularizing the inter-domain variance of risks with V-REx. The 
global minimum is now much closer to the top right section, which ensures the loss is also minimized on this domain.
The bottom left section is now repulsive to avoid overfitting and only converging to parameter values with approximate 
equality of risks across domains. The minimum is now situated outside of the convex hull of training domains, which enables generalization by risk extrapolation as demonstrated in \cite{krueger2021out}.

\subsection{Robust Continuous Control on Distracting Control Suite}
We now evaluate VIBR on a set of tasks from the Distracting Control Suite benchmark \citep{stone2021distracting}. We measure 
training efficiency, robustness to distractions and out-of-distribution generalization capacities. We showcase aggregated
and detailed results per task and evaluation metrics, as well as detailed ablations and discussion on different components
of the loss.\

\paragraph{Baselines}\label{exp:baselines}\ 
We use VIBR on top of Soft-Actor Critic \citep{haarnoja2018soft} for continuous control in the DCS environment. Our implementation follows DrQ \citep{yarats2020image} and we compare ourselves with 4 other baselines learning 
view-invariant representations:
\begin{itemize}
  \item DBC \citep{zhang2021learning}: a metric-based self-supervised learning objective that only keep task-relevant 
features in the representation
  \item SPR \citep{schwarzer2021data}: a self-supervised next latent state prediction auxiliary task 
  \item CURL \citep{laskin2020curl}: a contrastive learning objective inspired from computer vision
  \item Feature-Matching: a simple baseline where we match encoder outputs between different views
  \item DrQ \citep{yarats2020image}: a model-free baseline with no representation learning auxiliary loss
\end{itemize}
We detail each loss in Appendix \ref{appendix:representation}. For each experience, we train 4 random seeds for over 500k steps of 
gradient descent with Adam optimizer. We use the SAC implementation of ACME \citep{hoffman2020acme} in Jax 
\citep{jax2018github} for faster training. 

\paragraph{Training and Evaluation.}\
With DCS, we create a curriculum of 5 evaluation domains with progressive difficulty ranging from the 
vanilla environment with no distractor (CO) to intense dynamic visual perturbations (C4) such as random camera movements,
color randomization and extreme background randomization. Evaluation domains distributions 
are purposefully nested inside each other, to properly evaluate for in-distribution and out-of-distribution generalization:
$C0 \subset C1 \subset C2 \subset C3 \subset C4$. Details about the implementation of the curriculum can be found in Appendix
\ref{appendix:DCS}. By construction, VIBR requires multiple observers to work. We choose $K=2$
observers to limit compute intensity, although the method can be applied with more. We show that results are already 
very strong with two observers. VIBR is trained with an observer in C0 and an observer in C2. \textbf{Importantly, all baselines are trained with the same data as VIBR, with access to both C0 and C2 at every timestep.} Specifically, the baselines use their online-target architectures to pass each view through a different branch of the network, in the spirit of VIBR or more broadly self-supervised learning methodologies in computer vision. More details about implementation can be found 
in Appendix \ref{appendix:implementation}.\

We evaluate methods every 50k steps and at the end of training and accumulate the return over each episode. 
Episodic return are normalized to 1. We systematically use the \textit{rlliable} library \citep{agarwal2021deep} 
to evaluate our models, using stratified bootstrap over seeds and/or tasks on the benchmark to provide 
robust evaluation metrics. 
In particular, we use the inter-quartile mean (IQM) as a robust replacement to the mean 
while being more sample efficient than the median. We also use define a \textit{generalization gap} metric: 
\begin{equation}
  \mathcal{G}(C_i)=1 - \frac{\mathrm{IQM}(C_i)}{\mathrm{IQM}(C0)}
\end{equation}
This measures allows us to measure the drop in performance purely caused by domain shift uncorrelated from potential 
sub-optimal training. To properly test for generalization, evaluation environments use a different dataset 
of videos for the background even when training and evaluation have the same distraction difficulty. As such, we 
specifically refer to training domains as C1*, C2* and C3* to mark the point.

\subsubsection{Evaluation Results on Distracting Control}

\begin{figure}[t!]
  \centering
  \begin{subfigure}[t]{0.49\textwidth}
      \centering
      \includegraphics[width=\textwidth]{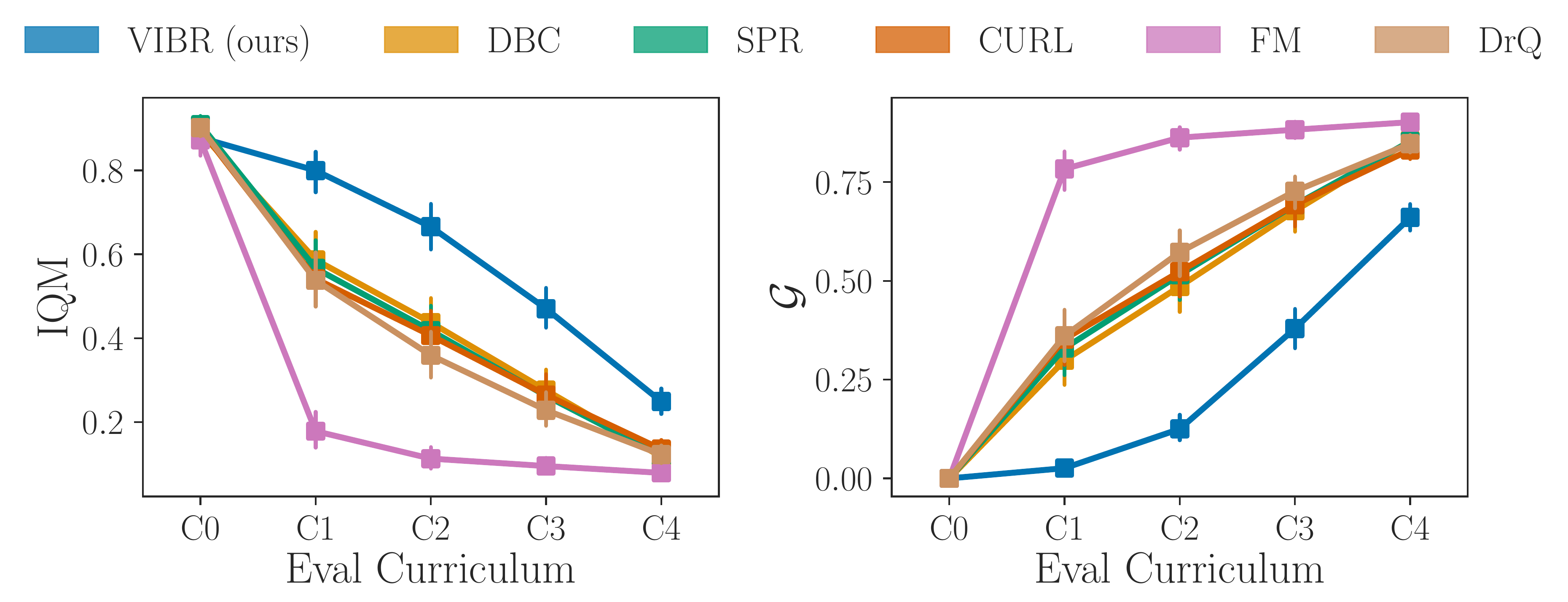}
      \caption{}
      \label{fig:main_eval}
  \end{subfigure}
  \hfill
  \begin{subfigure}[t]{0.49\textwidth}
      \centering
      \includegraphics[width=\textwidth]{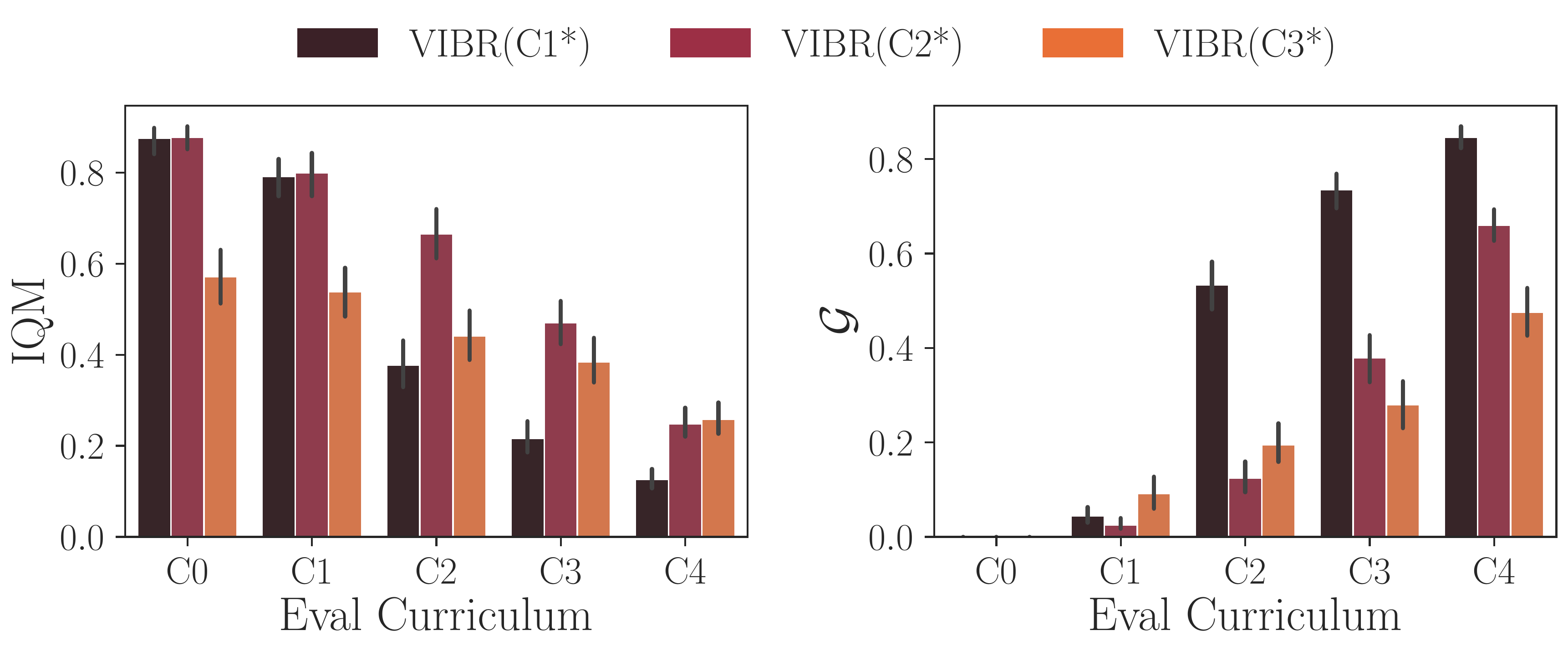}
      \caption{}
      \label{fig:train_eval}
  \end{subfigure}
 \caption{\textbf{(a):} Evaluation score (IQM and Generalization Gap) of VIBR and baselines over all 5 evaluation domains. Vertical bars are bootstrapped CI. \textbf{(b):} Effect of training curriculum on generalization.}
 \label{fig:ful_eval}
\end{figure}

\begin{figure}[t!]
  \centering
  \begin{subfigure}[t]{0.33\textwidth}
      \centering
      \includegraphics[width=\textwidth]{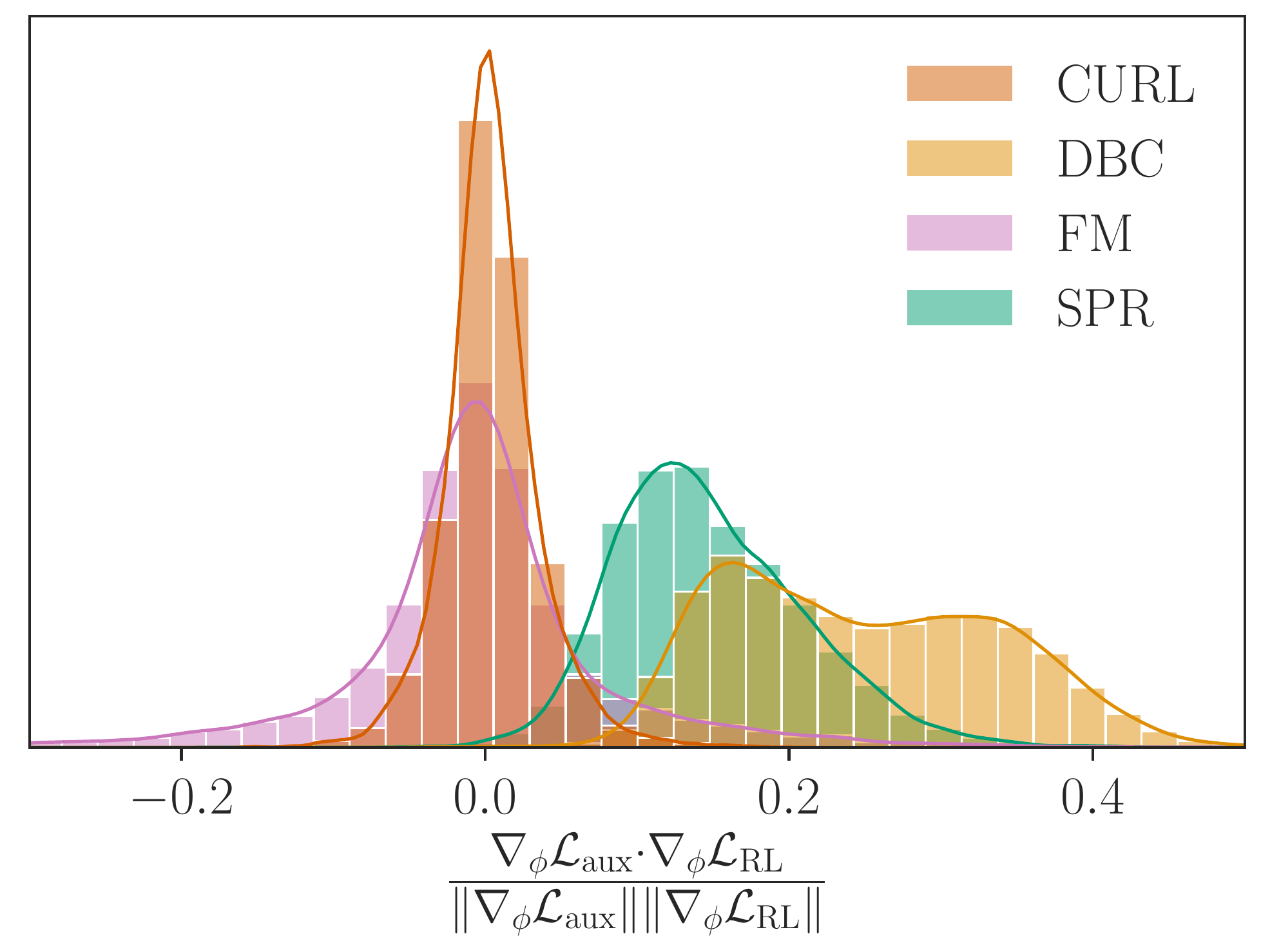}
      \caption{}
      \label{fig:cosine}
  \end{subfigure}
  \hfill
  \begin{subfigure}[t]{0.33\textwidth}
      \centering
      \includegraphics[width=\textwidth]{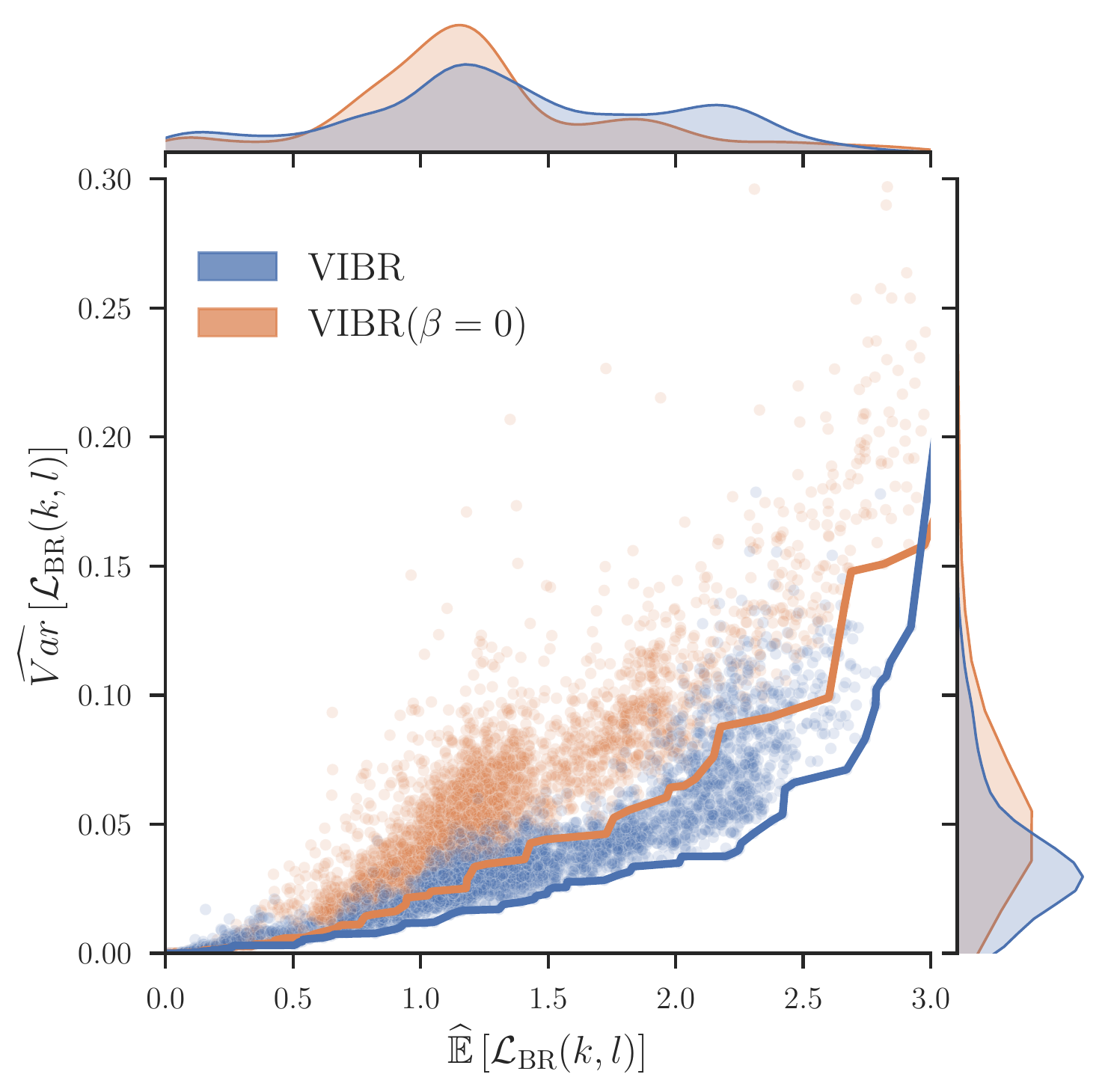}
      \caption{}
      \label{fig:jointplot}
  \end{subfigure}
  \hfill
  \begin{subfigure}[t]{0.33\textwidth}
      \centering
      \includegraphics[width=\textwidth]{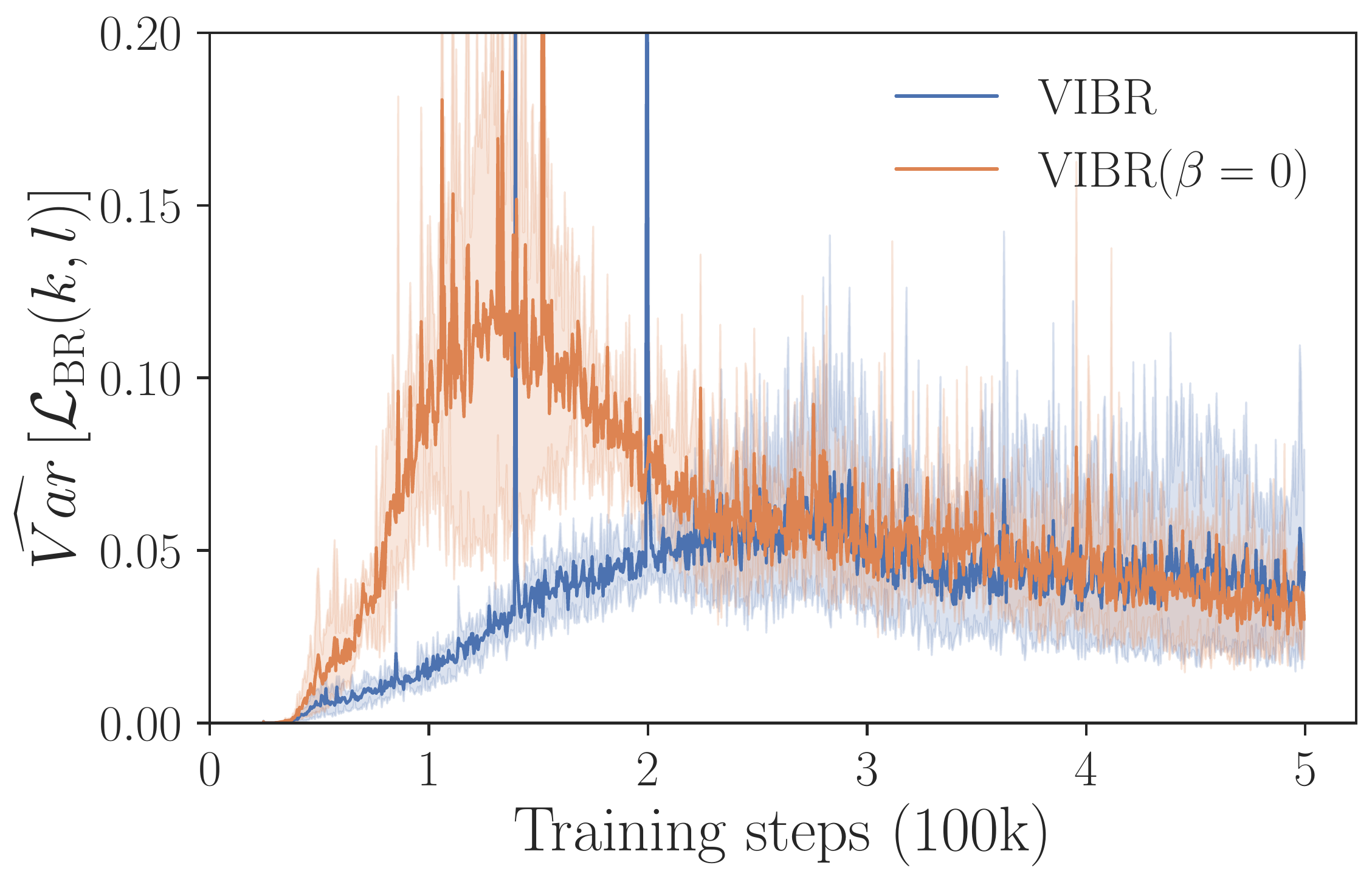}
      \caption{}
      \label{fig:var_q_loss}
  \end{subfigure}
 \caption{\textbf{(a):} Cosine similarity between RL and auxiliary task during training of representation learninig baselines. \textbf{(b):} Distribution and lower Pareto frontier of VIBR loss components during training over 4 seeds.  \textbf{(c):} Evolution of empirical inter-observer variance loss during training.}
 \label{fig:training_analysis}
\end{figure}

\paragraph{Aggregated performance} In Figure \ref{fig:iqm}, we aggregate IQM and generalization gap across C1 $\cdots$ C4 and
show the results for VIBR as well as the baselines described in \ref{exp:baselines}. VIBR improves IQM by 
65 \% and reduces generalization gap by 54 \% over the best performing baseline DBC. While CURL, SPR and DBC performed
similarly, FM is the only representation learning baselines to completely fail the task. Although the pretext task is quite
similar the other pretext tasks, FM lacks a projector network in its teacher-student architecture which is known to 
help performance by preventing the pretext task from directly optimizing on the encoder 
\citep{grill2020bootstrap, chen2020improved, bardes2022vicreg} and alleviating gradient conflicts. Yet, neither of the 
representation learning baselines reach the performance of VIBR which has less parameters and a simpler objective. 
We hypothesize that gradient conflicts between the auxiliary and RL task might explain the drop in performance. We
empirically validate this hypothesis in Figure \ref{fig:cosine} where we plot the whole distribution of cosine similarity
between the auxiliary loss and the RL loss during training. Overall, all four methods show weak gradient alignment 
with the RL objective. We notice however a positive correlation between IQM/generalization gap performance and 
average cosine similarity. Methods like DBC and SPR which rank higher also show non-zero average cosine-similarity during
training. On the other hand, CURL and FM showcase Gaussian distribution centered around zero.
Interestingly, DrQ is already a simple yet very strong baseline. This reflects that on hard optimization problems with 
many distractions like DCS (and arguably the real world), 
purely data-based end-to-end methods with carefully selected objective might be more efficient than intermediate 
methodological improvements on representation learning.

\begin{wrapfigure}{r}{0.3\textwidth}
  \begin{center}
    \includegraphics[width=.3\textwidth]{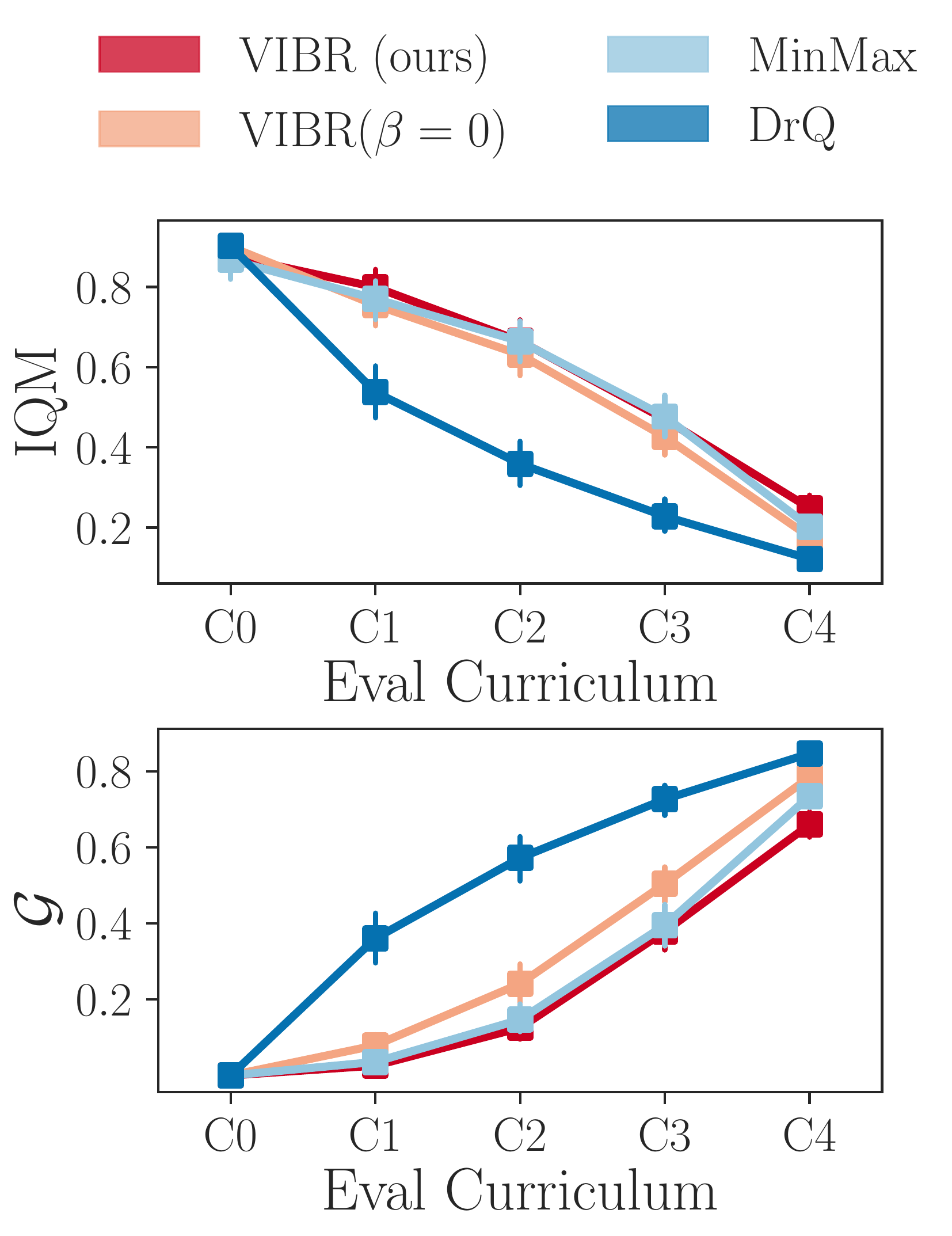}
  \end{center}
  \caption{Evaluation score over ablations and variations of VIBR. Shaded areas are bootstrapped CI.}
  \label{fig:additional}
\end{wrapfigure}

\paragraph{Detailed Results on the Evaluation Curriculum} Figure \ref{fig:main_eval} show the comparison of VIBR with 
the baselines in details across all evaluation domains. All methods were trained using C0* and C2* as defined above.
VIBR and all baselines have the same performance on C0
(without perturbations, identical to one of the training domain), which shows that aggregated score differences cannot 
be explained by difficulties with learning the control policy in a clean setting. 
Representation learning methods (except FM) show slight improvement
of generalization over DrQ with a better IQM and generalization gap on C2 and C3. However, none of the baseline is able
to achieve a statistically significant progress on C4, the most challenging benchmark. VIBR improves IQM and generalization 
gap on all benchmarks from C1 to C4, while keeping competitive performance on C0. Not only did it learn good control policies,
which we evaluate with C0, but it also developed interpolation and extrapolation capacities with a large increase 
both in-distribution (C1 and C2) and out-of-distribution domains (C3 and C4).\

\paragraph{In-distribution vs Out-of-distribution Generalization} Next, we evaluate in Figure \ref{fig:train_eval} how does VIBR distributes model capacity and extrapolates across domains when training benchmarks are 
in the form $C0 + Ck*$ with $k \in \left[ 1,3 \right]$. This allows us to modulate which benchmark are 
in the interpolation or extrapolation regime
in $\mathcal{O}$ (as depicted in orange and blue respectively in Figure \ref{fig:schematics}).

We observe a flattening of the performance curve as we transition from $C1*$ to $C3*$ as the secondary training domain. VIBR(C2*) shows improved IQM over VIBR(C1*) on C2,C3 and C4, as well as significant decrease in generalization gap on all generalization benchmarks. Overall, this translates into a pure 
increase of aggregated performance which means the model is able to distribute better its capacity over the image space while still functioning well in no-distraction regions. VIBR(C3*) however loses IQM over VIBR(C2*) in all but C4 which largely flattens the IQM over domains but keeps significantly decreasing the generalization gap on C3 and C4. This demonstrates that although VIBR(C3*) is seeing optimization difficulties (drop in C0 performance), training with more visual diversity (C3*) still keeps on improving OOD generalization (C4). Training on harder domains helps generalization to harder benchmarks as expected, but reduce overall performance: the network capacity and training time remained constant while the training task became harder.\

\paragraph{Impact of Multi-View Training}

We compare VIBR with tuned $\beta$ with an ablation where $\beta=0$. This recovers the setting of ERM if we consider cross-domain terms of the form $\mathcal{L}_{BR}(k,l)$ with $k \neq l$ as each a single training pseudo-domain. Figure \ref{fig:additional} show that these terms already have a big impact on performance compared to the DrQ baselines, which simply performs crop-resize augmentation with both actor and critic losses averaged over the two real training domains (C0 and C2). We also compare with \textbf{Minmax}, a variant of our objective that performs robust optimization instead of risk invariance by minimzing the worst-case risk over all Bellman residuals instead of the average: 
    $\theta \in \arg \min_\theta \max_{(k,l)} \mathcal{L}_{BR}(k,l)$.
This variant almost matches the performance of VIBR but remains slightly below. \cite{krueger2021out} proved the connection between V-REx and robust optimization and showed that V-REx has slightly better gradients, which can explain the small yet existing performance gap in our experiment.

\paragraph{Influence of the Risk Extrapolation Term}

We first investigate how does the variance regularization term weighted by $\beta$ influences learning dynamics and help generalization.
During training, we save the pair $\left( \widehat{\mathbb{E}}\left[ \mathcal{L}_\mathrm{BR}(k,l)\right], \widehat{Var}\left[ \mathcal{L} \mathrm{BR}(k,l)\right] \right)$ for every batch and plot them in Figure \ref{fig:jointplot}. We compare training with $\beta >0$ and 
$\beta=0$ on Walker Walk aggregated on 4 seeds. As visible by the lower Pareto frontier and the marginals, the regularizer has the intended
effect described in section~\ref{meth:OOD}: a positive $\beta$ shifts the overall distribution towards lower $\widehat{Var}\left[ \mathcal{L}
\mathrm{BR}(k,l)\right]$ during training. Consequently, the marginal $\widehat{\mathbb{E}}\left[ \mathcal{L}_\mathrm{BR}(k,l)\right]$ is more uniform and less concentrated around lower values when $\beta>O$. This means that the model effectively perform a trade-off between
bias on some of the views/observers in order to keep the variance low. When we plot the distribution of $\widehat{Var}\left[ \mathcal{L}_\mathrm{BR}(k,l)\right]$ across time steps (Figure \ref{fig:var_q_loss}), we notice that most of the regularization is impactful at the beginning of training but does not affect asymptotic convergence. This mechanism is particularly helpful as deep reinforcement learning networks are known to suffer from early overfitting preventing them from reaching higher performance in the long run \citep{nikishin2022primacy}. In our case, early overfitting happens on the training domain C0* which is indicated by excellent performance of all baselines on C0 even when generalization fails.

\paragraph{Additional Results} We study the $\beta$ hyperparamter by measuring IQM as a function of $\beta$ for all 6 environments of the DCS (Figure~\ref{fig:beta}). Notably, all environments approximately converge to a zero variance of IQM between evaluation domains. We notice that difference in environment dynamics influence the choice for optimal $\beta$. Note that we did not change the default action repeat hyperparameter of DCS. This makes Cartpole Swingup an extremely difficult task with an action repeat of 8 while having dynamic distractions in the background and explains the surprisingly low performance despite the simple mechanics. Further detailed results can be found in Appendix \ref{appendix:efficiency} and \ref{appendix:iqm}, as well as an ablation study of the loss components in Appendix \ref{appendix:ablation} showing the importance of using all pairwise bellman residuals in VIBR.

\section{Related Works}

\paragraph{Robust Visual RL}
Although not studied in this work, data augmentation is an efficient and easy to implement method to regularize models and increase generalization performance. Some methods study which type of data augmentation is better suited for RL: random shifts \citep{yarats2020image, yarats2021mastering, laskin2020reinforcement}, data mixing and convolutions \citep{wang2020improving, zhang2021generalization, zhou2020domain} or masking \citep{lee2020network, seo2023masked, xiao2022masked}. A more efficient method for robust RL  is using domain randomization. \cite{hansen2021generalization, stone2021distracting, grigsby2020measuring} modify the original Deepmind Control Suite 
\citep{tassa2018deepmind} to include visual distractions, 
\cite{xing2021kitchenshift, zhu2020robosuite, ahmed2020causalworld} involve robotic tasks and and causality.  \cite{akkaya2019solving} dynamically adapts domain randomization intensity and
\cite{ren2020improving} use an adversarial objective. Other works use parallel environments with different randomizations \citep{ren2020improving, li2021domain, zhao2021robust, james2019sim, zhang2020invariant}.\

\paragraph{Invariant Representation Learning}
 is another approach to ensure good generalization across visual perturbations.
 \cite{zhang2021learning, agarwal2021contrastive, bertran2022efficient} uses behavioural metric learning. Other works relate invariant prediction \citep{peters2016causal, arjovsky2019invariant}, robust optimization and causal inference to isolate causal feature sets and keep only task-relevant features , \citep{zhang2020invariant, sonar2021invariant}. Multiple work use the Block-MDP setting to learn invariant representations \citep{zhanglearning, zhang2021learning, agarwal2021contrastive, bertran2022efficient, efroni2021provably}. Other works focus on model learning: \cite{lu2020sample} combines data augmentation with counterfactuals to learn a structured 
 causal model with an adversarial objective, while \cite{wang2022denoised} learn noise-invariant world-models. \cite{li2021domain} also uses an adversarial objective combined with gradient reversal to learn a representation robust to interventions.
 \cite{mozifian2020intervention} apply bisimulation and risk extrapolation \citep{krueger2021out} on robotics. We differ from this line of work as VIBR  does not need any auxiliary representation learning loss.


\section{Discussion and conclusion}

Our method has the obvious limitation of requiring multiple views during training. However, we emphasize that our multi-view assumption is only necessary during the learning phase and not at inference time. This becomes particularly advantageous in Sim2Real settings, where simulating multiple points of view is cost-effective, whereas providing multiple viewpoints at inference may be sometimes more challenging. Simulations can be used to intervene on the visual aspects of the environment without necessitating additional views for real-world inference. \

However, as there is a limited number of simulated benchmarks specifically designed for multi-view training setups in image-based reinforcement learning, it is both plausible and practical to use multiple views at inference time in many real-life scenarios using multiple cameras. This is an affordable and prevalent practice in robotics platforms and autonomous vehicles, which often employ multiple sensors and cameras to address redundancies and potential occlusions. \

While the baselines do not require this assumption, limiting the training setup to a single frame or view of the scene can therefore be unnecessarily restrictive and may not reflect actual practical use cases. The main objective of this paper is to serve as a proof-of-concept for the efficient utilization of existing multi-view training setups, such as those already found in the robotics domain. \

Therefore, assuming multiple view availability, we presented VIBR (View-Invariant Bellman Residuals), a method that combines multi-view training and invariant prediction to reduce out-of-distribution (OOD) generalization gap for RL based visuomotor control. We demonstrated strong generalization results on the Distracting Control Suite benchmark and to our knowledge, VIBR is the first method to provide non-trivial performance on the hardest setting of DCS (C4). We also provided an analysis of the learning dynamics of VIBR which helped us explain its competitive performance compared to representation learning methods. Further work include finding appropriate architectures for multi-view training, scaling the model and data size to tackle more complex visuomotor control challenges, automatic tuning of $\beta$ with (meta-)learning or heuristics and leveraging pretrained generative models to sample observers without having access to a simulation or multiple cameras. 

\subsubsection*{Acknowledgments}
This publication was made possible by the use of the FactoryIA supercomputer, financially supported by the Ile-De-France Regional Council.

\bibliography{collas2023_conference}

\begin{thebibliography}{49}
\providecommand{\natexlab}[1]{#1}
\providecommand{\url}[1]{\texttt{#1}}
\expandafter\ifx\csname urlstyle\endcsname\relax
  \providecommand{\doi}[1]{doi: #1}\else
  \providecommand{\doi}{doi: \begingroup \urlstyle{rm}\Url}\fi

\bibitem[Agarwal et~al.(2021{\natexlab{a}})Agarwal, Machado, Castro, and
  Bellemare]{agarwal2021contrastive}
Rishabh Agarwal, Marlos~C Machado, Pablo~Samuel Castro, and Marc~G Bellemare.
\newblock Contrastive behavioral similarity embeddings for generalization in
  reinforcement learning.
\newblock In \emph{International Conference on Learning Representations},
  2021{\natexlab{a}}.

\bibitem[Agarwal et~al.(2021{\natexlab{b}})Agarwal, Schwarzer, Castro,
  Courville, and Bellemare]{agarwal2021deep}
Rishabh Agarwal, Max Schwarzer, Pablo~Samuel Castro, Aaron~C Courville, and
  Marc Bellemare.
\newblock Deep reinforcement learning at the edge of the statistical precipice.
\newblock \emph{Advances in Neural Information Processing Systems}, 34,
  2021{\natexlab{b}}.

\bibitem[Ahmed et~al.(2020)Ahmed, Tr{\"a}uble, Goyal, Neitz, Wuthrich, Bengio,
  Sch{\"o}lkopf, and Bauer]{ahmed2020causalworld}
Ossama Ahmed, Frederik Tr{\"a}uble, Anirudh Goyal, Alexander Neitz, Manuel
  Wuthrich, Yoshua Bengio, Bernhard Sch{\"o}lkopf, and Stefan Bauer.
\newblock Causalworld: A robotic manipulation benchmark for causal structure
  and transfer learning.
\newblock In \emph{International Conference on Learning Representations}, 2020.

\bibitem[Akkaya et~al.(2019)Akkaya, Andrychowicz, Chociej, Litwin, McGrew,
  Petron, Paino, Plappert, Powell, Ribas, et~al.]{akkaya2019solving}
Ilge Akkaya, Marcin Andrychowicz, Maciek Chociej, Mateusz Litwin, Bob McGrew,
  Arthur Petron, Alex Paino, Matthias Plappert, Glenn Powell, Raphael Ribas,
  et~al.
\newblock Solving rubik's cube with a robot hand.
\newblock \emph{arXiv preprint arXiv:1910.07113}, 2019.

\bibitem[Arjovsky et~al.(2019)Arjovsky, Bottou, Gulrajani, and
  Lopez-Paz]{arjovsky2019invariant}
Martin Arjovsky, L{\'e}on Bottou, Ishaan Gulrajani, and David Lopez-Paz.
\newblock Invariant risk minimization.
\newblock \emph{arXiv preprint arXiv:1907.02893}, 2019.

\bibitem[Bardes et~al.(2022)Bardes, Ponce, and Lecun]{bardes2022vicreg}
Adrien Bardes, Jean Ponce, and Yann Lecun.
\newblock Vicreg: Variance-invariance-covariance regularization for
  self-supervised learning.
\newblock In \emph{ICLR 2022-10th International Conference on Learning
  Representations}, 2022.

\bibitem[Bellemare et~al.(2019)Bellemare, Dabney, Dadashi, Ali~Taiga, Castro,
  Le~Roux, Schuurmans, Lattimore, and Lyle]{bellemare2019geometric}
Marc Bellemare, Will Dabney, Robert Dadashi, Adrien Ali~Taiga, Pablo~Samuel
  Castro, Nicolas Le~Roux, Dale Schuurmans, Tor Lattimore, and Clare Lyle.
\newblock A geometric perspective on optimal representations for reinforcement
  learning.
\newblock \emph{Advances in neural information processing systems}, 32, 2019.

\bibitem[Bertran et~al.(2022)Bertran, Talbott, Srivastava, and
  Susskind]{bertran2022efficient}
Martin Bertran, Walter Talbott, Nitish Srivastava, and Joshua Susskind.
\newblock Efficient embedding of semantic similarity in control policies via
  entangled bisimulation.
\newblock \emph{arXiv preprint arXiv:2201.12300}, 2022.

\bibitem[Bradbury et~al.(2018)Bradbury, Frostig, Hawkins, Johnson, Leary,
  Maclaurin, Necula, Paszke, Vander{P}las, Wanderman-{M}ilne, and
  Zhang]{jax2018github}
James Bradbury, Roy Frostig, Peter Hawkins, Matthew~James Johnson, Chris Leary,
  Dougal Maclaurin, George Necula, Adam Paszke, Jake Vander{P}las, Skye
  Wanderman-{M}ilne, and Qiao Zhang.
\newblock {JAX}: composable transformations of {P}ython+{N}um{P}y programs,
  2018.
\newblock URL \url{http://github.com/google/jax}.

\bibitem[Chen et~al.(2020)Chen, Fan, Girshick, and He]{chen2020improved}
Xinlei Chen, Haoqi Fan, Ross Girshick, and Kaiming He.
\newblock Improved baselines with momentum contrastive learning.
\newblock \emph{arXiv preprint arXiv:2003.04297}, 2020.

\bibitem[Dabney et~al.(2021)Dabney, Barreto, Rowland, Dadashi, Quan, Bellemare,
  and Silver]{dabney2021value}
Will Dabney, Andr{\'e} Barreto, Mark Rowland, Robert Dadashi, John Quan, Marc~G
  Bellemare, and David Silver.
\newblock The value-improvement path: Towards better representations for
  reinforcement learning.
\newblock In \emph{Proceedings of the AAAI Conference on Artificial
  Intelligence}, volume~35, pp.\  7160--7168, 2021.

\bibitem[Du et~al.(2019)Du, Krishnamurthy, Jiang, Agarwal, Dudik, and
  Langford]{du2019provably}
Simon Du, Akshay Krishnamurthy, Nan Jiang, Alekh Agarwal, Miroslav Dudik, and
  John Langford.
\newblock Provably efficient rl with rich observations via latent state
  decoding.
\newblock In \emph{International Conference on Machine Learning}, pp.\
  1665--1674. PMLR, 2019.

\bibitem[Efroni et~al.(2021)Efroni, Misra, Krishnamurthy, Agarwal, and
  Langford]{efroni2021provably}
Yonathan Efroni, Dipendra Misra, Akshay Krishnamurthy, Alekh Agarwal, and John
  Langford.
\newblock Provably filtering exogenous distractors using multistep inverse
  dynamics.
\newblock In \emph{International Conference on Learning Representations}, 2021.

\bibitem[Grigsby \& Qi(2020)Grigsby and Qi]{grigsby2020measuring}
Jake Grigsby and Yanjun Qi.
\newblock Measuring visual generalization in continuous control from pixels.
\newblock \emph{arXiv preprint arXiv:2010.06740}, 2020.

\bibitem[Grill et~al.(2020)Grill, Strub, Altch{\'e}, Tallec, Richemond,
  Buchatskaya, Doersch, Avila~Pires, Guo, Gheshlaghi~Azar,
  et~al.]{grill2020bootstrap}
Jean-Bastien Grill, Florian Strub, Florent Altch{\'e}, Corentin Tallec, Pierre
  Richemond, Elena Buchatskaya, Carl Doersch, Bernardo Avila~Pires, Zhaohan
  Guo, Mohammad Gheshlaghi~Azar, et~al.
\newblock Bootstrap your own latent-a new approach to self-supervised learning.
\newblock \emph{Advances in neural information processing systems},
  33:\penalty0 21271--21284, 2020.

\bibitem[Haarnoja et~al.(2018)Haarnoja, Zhou, Abbeel, and
  Levine]{haarnoja2018soft}
Tuomas Haarnoja, Aurick Zhou, Pieter Abbeel, and Sergey Levine.
\newblock Soft actor-critic: Off-policy maximum entropy deep reinforcement
  learning with a stochastic actor.
\newblock In \emph{International conference on machine learning}, pp.\
  1861--1870. PMLR, 2018.

\bibitem[Hansen \& Wang(2021)Hansen and Wang]{hansen2021generalization}
Nicklas Hansen and Xiaolong Wang.
\newblock Generalization in reinforcement learning by soft data augmentation.
\newblock In \emph{2021 IEEE International Conference on Robotics and
  Automation (ICRA)}, pp.\  13611--13617. IEEE, 2021.

\bibitem[Hoffman et~al.(2020)Hoffman, Shahriari, Aslanides, Barth-Maron,
  Momchev, Sinopalnikov, Sta\'nczyk, Ramos, Raichuk, Vincent, Hussenot,
  Dadashi, Dulac-Arnold, Orsini, Jacq, Ferret, Vieillard, Ghasemipour, Girgin,
  Pietquin, Behbahani, Norman, Abdolmaleki, Cassirer, Yang, Baumli, Henderson,
  Friesen, Haroun, Novikov, Colmenarejo, Cabi, Gulcehre, Paine, Srinivasan,
  Cowie, Wang, Piot, and de~Freitas]{hoffman2020acme}
Matthew~W. Hoffman, Bobak Shahriari, John Aslanides, Gabriel Barth-Maron,
  Nikola Momchev, Danila Sinopalnikov, Piotr Sta\'nczyk, Sabela Ramos, Anton
  Raichuk, Damien Vincent, L\'eonard Hussenot, Robert Dadashi, Gabriel
  Dulac-Arnold, Manu Orsini, Alexis Jacq, Johan Ferret, Nino Vieillard, Seyed
  Kamyar~Seyed Ghasemipour, Sertan Girgin, Olivier Pietquin, Feryal Behbahani,
  Tamara Norman, Abbas Abdolmaleki, Albin Cassirer, Fan Yang, Kate Baumli,
  Sarah Henderson, Abe Friesen, Ruba Haroun, Alex Novikov, Sergio~G\'omez
  Colmenarejo, Serkan Cabi, Caglar Gulcehre, Tom~Le Paine, Srivatsan
  Srinivasan, Andrew Cowie, Ziyu Wang, Bilal Piot, and Nando de~Freitas.
\newblock Acme: A research framework for distributed reinforcement learning.
\newblock \emph{arXiv preprint arXiv:2006.00979}, 2020.

\bibitem[Jaderberg et~al.(2017)Jaderberg, Mnih, Czarnecki, Schaul, Leibo,
  Silver, and Kavukcuoglu]{jaderbergreinforcement}
Max Jaderberg, Volodymyr Mnih, Wojciech~Marian Czarnecki, Tom Schaul, Joel~Z
  Leibo, David Silver, and Koray Kavukcuoglu.
\newblock Reinforcement learning with unsupervised auxiliary tasks.
\newblock In \emph{International Conference on Learning Representations}, 2017.

\bibitem[James et~al.(2019)James, Wohlhart, Kalakrishnan, Kalashnikov, Irpan,
  Ibarz, Levine, Hadsell, and Bousmalis]{james2019sim}
Stephen James, Paul Wohlhart, Mrinal Kalakrishnan, Dmitry Kalashnikov, Alex
  Irpan, Julian Ibarz, Sergey Levine, Raia Hadsell, and Konstantinos Bousmalis.
\newblock Sim-to-real via sim-to-sim: Data-efficient robotic grasping via
  randomized-to-canonical adaptation networks.
\newblock In \emph{Proceedings of the IEEE/CVF Conference on Computer Vision
  and Pattern Recognition}, pp.\  12627--12637, 2019.

\bibitem[Krueger et~al.(2021)Krueger, Caballero, Jacobsen, Zhang, Binas, Zhang,
  Le~Priol, and Courville]{krueger2021out}
David Krueger, Ethan Caballero, Joern-Henrik Jacobsen, Amy Zhang, Jonathan
  Binas, Dinghuai Zhang, Remi Le~Priol, and Aaron Courville.
\newblock Out-of-distribution generalization via risk extrapolation (rex).
\newblock In \emph{International Conference on Machine Learning}, pp.\
  5815--5826. PMLR, 2021.

\bibitem[Laskin et~al.(2020{\natexlab{a}})Laskin, Srinivas, and
  Abbeel]{laskin2020curl}
Michael Laskin, Aravind Srinivas, and Pieter Abbeel.
\newblock Curl: Contrastive unsupervised representations for reinforcement
  learning.
\newblock In \emph{International Conference on Machine Learning}, pp.\
  5639--5650. PMLR, 2020{\natexlab{a}}.

\bibitem[Laskin et~al.(2020{\natexlab{b}})Laskin, Lee, Stooke, Pinto, Abbeel,
  and Srinivas]{laskin2020reinforcement}
Misha Laskin, Kimin Lee, Adam Stooke, Lerrel Pinto, Pieter Abbeel, and Aravind
  Srinivas.
\newblock Reinforcement learning with augmented data.
\newblock \emph{Advances in neural information processing systems},
  33:\penalty0 19884--19895, 2020{\natexlab{b}}.

\bibitem[Lee et~al.(2020)Lee, Lee, Shin, and Lee]{lee2020network}
Kimin Lee, Kibok Lee, Jinwoo Shin, and Honglak Lee.
\newblock Network randomization: A simple technique for generalization in deep
  reinforcement learning.
\newblock In \emph{International Conference on Learning Representations}, 2020.

\bibitem[Li et~al.(2021)Li, Fran{\c{c}}ois-Lavet, Doan, and
  Pineau]{li2021domain}
Bonnie Li, Vincent Fran{\c{c}}ois-Lavet, Thang Doan, and Joelle Pineau.
\newblock Domain adversarial reinforcement learning.
\newblock \emph{arXiv preprint arXiv:2102.07097}, 2021.

\bibitem[Lu et~al.(2020)Lu, Huang, Wang, Hern{\'a}ndez-Lobato, Zhang, and
  Sch{\"o}lkopf]{lu2020sample}
Chaochao Lu, Biwei Huang, Ke~Wang, Jos{\'e}~Miguel Hern{\'a}ndez-Lobato, Kun
  Zhang, and Bernhard Sch{\"o}lkopf.
\newblock Sample-efficient reinforcement learning via counterfactual-based data
  augmentation.
\newblock \emph{arXiv preprint arXiv:2012.09092}, 2020.

\bibitem[Mozifian et~al.(2020)Mozifian, Zhang, Pineau, and
  Meger]{mozifian2020intervention}
Melissa Mozifian, Amy Zhang, Joelle Pineau, and David Meger.
\newblock Intervention design for effective sim2real transfer.
\newblock \emph{arXiv preprint arXiv:2012.02055}, 2020.

\bibitem[Nikishin et~al.(2022)Nikishin, Schwarzer, D’Oro, Bacon, and
  Courville]{nikishin2022primacy}
Evgenii Nikishin, Max Schwarzer, Pierluca D’Oro, Pierre-Luc Bacon, and Aaron
  Courville.
\newblock The primacy bias in deep reinforcement learning.
\newblock In \emph{International Conference on Machine Learning}, pp.\
  16828--16847. PMLR, 2022.

\bibitem[Peters et~al.(2016)Peters, B{\"u}hlmann, and
  Meinshausen]{peters2016causal}
Jonas Peters, Peter B{\"u}hlmann, and Nicolai Meinshausen.
\newblock Causal inference by using invariant prediction: identification and
  confidence intervals.
\newblock \emph{Journal of the Royal Statistical Society: Series B (Statistical
  Methodology)}, 78\penalty0 (5):\penalty0 947--1012, 2016.

\bibitem[Rahaman et~al.(2019)Rahaman, Baratin, Arpit, Draxler, Lin, Hamprecht,
  Bengio, and Courville]{rahaman2019spectral}
Nasim Rahaman, Aristide Baratin, Devansh Arpit, Felix Draxler, Min Lin, Fred
  Hamprecht, Yoshua Bengio, and Aaron Courville.
\newblock On the spectral bias of neural networks.
\newblock In \emph{International Conference on Machine Learning}, pp.\
  5301--5310. PMLR, 2019.

\bibitem[Ren et~al.(2020)Ren, Duan, Li, Guan, and Sun]{ren2020improving}
Yangang Ren, Jingliang Duan, Shengbo~Eben Li, Yang Guan, and Qi~Sun.
\newblock Improving generalization of reinforcement learning with minimax
  distributional soft actor-critic.
\newblock In \emph{2020 IEEE 23rd International Conference on Intelligent
  Transportation Systems (ITSC)}, pp.\  1--6. IEEE, 2020.

\bibitem[Schwarzer et~al.(2021)Schwarzer, Anand, Goel, Hjelm, Courville, and
  Bachman]{schwarzer2021data}
Max Schwarzer, Ankesh Anand, Rishab Goel, R~Devon Hjelm, Aaron Courville, and
  Philip Bachman.
\newblock Data-efficient reinforcement learning with self-predictive
  representations.
\newblock In \emph{International Conference on Learning Representations}, 2021.

\bibitem[Seo et~al.(2023)Seo, Hafner, Liu, Liu, James, Lee, and
  Abbeel]{seo2023masked}
Younggyo Seo, Danijar Hafner, Hao Liu, Fangchen Liu, Stephen James, Kimin Lee,
  and Pieter Abbeel.
\newblock Masked world models for visual control.
\newblock In \emph{Conference on Robot Learning}, pp.\  1332--1344. PMLR, 2023.

\bibitem[Sonar et~al.(2021)Sonar, Pacelli, and Majumdar]{sonar2021invariant}
Anoopkumar Sonar, Vincent Pacelli, and Anirudha Majumdar.
\newblock Invariant policy optimization: Towards stronger generalization in
  reinforcement learning.
\newblock In \emph{Learning for Dynamics and Control}, pp.\  21--33. PMLR,
  2021.

\bibitem[Stone et~al.(2021)Stone, Ramirez, Konolige, and
  Jonschkowski]{stone2021distracting}
Austin Stone, Oscar Ramirez, Kurt Konolige, and Rico Jonschkowski.
\newblock The distracting control suite--a challenging benchmark for
  reinforcement learning from pixels.
\newblock \emph{arXiv preprint arXiv:2101.02722}, 2021.

\bibitem[Tassa et~al.(2018)Tassa, Doron, Muldal, Erez, Li, Casas, Budden,
  Abdolmaleki, Merel, Lefrancq, et~al.]{tassa2018deepmind}
Yuval Tassa, Yotam Doron, Alistair Muldal, Tom Erez, Yazhe Li, Diego de~Las
  Casas, David Budden, Abbas Abdolmaleki, Josh Merel, Andrew Lefrancq, et~al.
\newblock Deepmind control suite.
\newblock \emph{arXiv preprint arXiv:1801.00690}, 2018.

\bibitem[Wang et~al.(2020)Wang, Kang, Shao, and Feng]{wang2020improving}
Kaixin Wang, Bingyi Kang, Jie Shao, and Jiashi Feng.
\newblock Improving generalization in reinforcement learning with mixture
  regularization.
\newblock \emph{Advances in Neural Information Processing Systems},
  33:\penalty0 7968--7978, 2020.

\bibitem[Wang et~al.(2022)Wang, Du, Torralba, Isola, Zhang, and
  Tian]{wang2022denoised}
Tongzhou Wang, Simon Du, Antonio Torralba, Phillip Isola, Amy Zhang, and
  Yuandong Tian.
\newblock Denoised mdps: Learning world models better than the world itself.
\newblock In \emph{International Conference on Machine Learning}, pp.\
  22591--22612. PMLR, 2022.

\bibitem[Xiao et~al.(2022)Xiao, Radosavovic, Darrell, and
  Malik]{xiao2022masked}
Tete Xiao, Ilija Radosavovic, Trevor Darrell, and Jitendra Malik.
\newblock Masked visual pre-training for motor control.
\newblock \emph{arXiv preprint arXiv:2203.06173}, 2022.

\bibitem[Xing et~al.(2021)Xing, Gupta, Powers, and Dean]{xing2021kitchenshift}
Eliot Xing, Abhinav Gupta, Sam Powers, and Victoria Dean.
\newblock Kitchenshift: Evaluating zero-shot generalization of imitation-based
  policy learning under domain shifts.
\newblock In \emph{NeurIPS 2021 Workshop on Distribution Shifts: Connecting
  Methods and Applications}, 2021.

\bibitem[Yarats et~al.(2020)Yarats, Kostrikov, and Fergus]{yarats2020image}
Denis Yarats, Ilya Kostrikov, and Rob Fergus.
\newblock Image augmentation is all you need: Regularizing deep reinforcement
  learning from pixels.
\newblock In \emph{International Conference on Learning Representations}, 2020.

\bibitem[Yarats et~al.(2021)Yarats, Fergus, Lazaric, and
  Pinto]{yarats2021mastering}
Denis Yarats, Rob Fergus, Alessandro Lazaric, and Lerrel Pinto.
\newblock Mastering visual continuous control: Improved data-augmented
  reinforcement learning.
\newblock In \emph{International Conference on Learning Representations}, 2021.

\bibitem[Zhang et~al.(2020)Zhang, Lyle, Sodhani, Filos, Kwiatkowska, Pineau,
  Gal, and Precup]{zhang2020invariant}
Amy Zhang, Clare Lyle, Shagun Sodhani, Angelos Filos, Marta Kwiatkowska, Joelle
  Pineau, Yarin Gal, and Doina Precup.
\newblock Invariant causal prediction for block mdps.
\newblock In \emph{International Conference on Machine Learning}, pp.\
  11214--11224. PMLR, 2020.

\bibitem[Zhang et~al.(2021{\natexlab{a}})Zhang, McAllister, Calandra, Gal, and
  Levine]{zhang2021learning}
Amy Zhang, Rowan~Thomas McAllister, Roberto Calandra, Yarin Gal, and Sergey
  Levine.
\newblock Learning invariant representations for reinforcement learning without
  reconstruction.
\newblock In \emph{International Conference on Learning Representations},
  2021{\natexlab{a}}.

\bibitem[Zhang et~al.(2021{\natexlab{b}})Zhang, Sodhani, Khetarpal, and
  Pineau]{zhanglearning}
Amy Zhang, Shagun Sodhani, Khimya Khetarpal, and Joelle Pineau.
\newblock Learning robust state abstractions for hidden-parameter block mdps.
\newblock In \emph{International Conference on Learning Representations},
  2021{\natexlab{b}}.

\bibitem[Zhang \& Guo(2021)Zhang and Guo]{zhang2021generalization}
Hanping Zhang and Yuhong Guo.
\newblock Generalization of reinforcement learning with policy-aware
  adversarial data augmentation.
\newblock \emph{arXiv preprint arXiv:2106.15587}, 2021.

\bibitem[Zhao \& Hospedales(2021)Zhao and Hospedales]{zhao2021robust}
Chenyang Zhao and Timothy Hospedales.
\newblock Robust domain randomised reinforcement learning through peer-to-peer
  distillation.
\newblock In \emph{Asian Conference on Machine Learning}, pp.\  1237--1252.
  PMLR, 2021.

\bibitem[Zhou et~al.(2020)Zhou, Yang, Qiao, and Xiang]{zhou2020domain}
Kaiyang Zhou, Yongxin Yang, Yu~Qiao, and Tao Xiang.
\newblock Domain generalization with mixstyle.
\newblock In \emph{International Conference on Learning Representations}, 2020.

\bibitem[Zhu et~al.(2020)Zhu, Wong, Mandlekar, and
  Mart{\'\i}n-Mart{\'\i}n]{zhu2020robosuite}
Yuke Zhu, Josiah Wong, Ajay Mandlekar, and Roberto Mart{\'\i}n-Mart{\'\i}n.
\newblock robosuite: A modular simulation framework and benchmark for robot
  learning.
\newblock \emph{arXiv preprint arXiv:2009.12293}, 2020.

\end{thebibliography}
\bibliographystyle{collas2023_conference}

\newpage

\appendix

\section{Impact of $\beta$ on VIBR gradient norm}
\label{appendix:gradient}

We show here that a strictly positive $\beta$ value for the VIBR loss helps during training by preventing overfitting towards a deep single domain minimum and encouraging convergence to moderate local minima, thus helping extrapolation.\ 
Let $X = \mathcal{L}_\mathrm{BR}(k,l)$. The gradients of VIBR loss can be written:
\begin{align*}
    \nabla \mathcal{L}_\mathrm{VIBR} &= \Hat{\mathbb{E}} \left[ \nabla X  \right] 
    + \beta \nabla \Hat{Var} (X)\\
        &=\Hat{\mathbb{E}} \left[ \nabla X  \right] 
    + \beta \nabla \left( \Hat{\mathbb{E}} \left[ X^2  \right] - \Hat{\mathbb{E}} \left[ X  \right]^2 \right) \\
        &= \Hat{\mathbb{E}} \left[ \nabla X  \right]
    + \beta \Hat{\mathbb{E}} \left[ 2 X \nabla X  \right]
    - 2 \beta \Hat{\mathbb{E}} \left[ \nabla X \right] \Hat{\mathbb{E}} \left[ X \right]\\
        &= \Hat{\mathbb{E}} \left[ \nabla X + 2 \beta \left( X \nabla X - \Hat{\mathbb{E}} \left[ X \right] \nabla X \right)  \right]\\
        &= \Hat{\mathbb{E}} \left[ \nabla X \left(1 + 2 \beta \left(X  - \Hat{\mathbb{E}} \left[ X \right] \right) \right)  \right]\\
\end{align*}
This proves that gradients of VIBR are completely aligned with TD-learning gradient when looking at a sample in particular, but only change the norm of the gradient based on the sign of $\left(X  - \Hat{\mathbb{E}} \left[ X \right] \right)$. As a result, for a given $(k,l)$ pair of observers:
\begin{itemize}
    \item if $\mathcal{L}_\mathrm{BR}(k,l) < \Hat{\mathbb{E}} \left[ {L}_\mathrm{BR} \right]$ (\textbf{overfitting} to one observer pair $(x^k, x^l)$, then the weight of this update is reduced $$\lVert \nabla \mathcal{L}_\mathrm{VIBR}(x^k,x^l)\rVert < \lVert \nabla {L}_\mathrm{BR}(k,l)\rVert$$
    \item if $\mathcal{L}_\mathrm{BR}(k,l) > \Hat{\mathbb{E}} \left[ {L}_\mathrm{BR} \right]$ (\textbf{underfitting} to one observer pair $(x^k, x^l)$, then the weight of this update is increased $$\lVert \nabla \mathcal{L}_\mathrm{VIBR}(x^k,x^l)\rVert > \lVert \nabla {L}_\mathrm{BR}(k,l)\rVert$$
\end{itemize}
This mechanism discourages overfitting to deep minimum and actively promotes converging towards "hard-to-optimize" regions. An example of such region is a domain that is located very far from the overfitting domain: ensuring good performance on it would be difficult as shown in Figure \ref{fig:loss_landscape}. Because OOD domains minima are more likely to be located far from obvious minimum where overfitting is frequent, the variance regularization term encourages fitting to domains that might be closer to OOD, which helps generalization.

\section{$\beta$ Hyperparameter Study}

\begin{figure}[h!]
  \begin{center}
    \includegraphics[width=.7\textwidth]{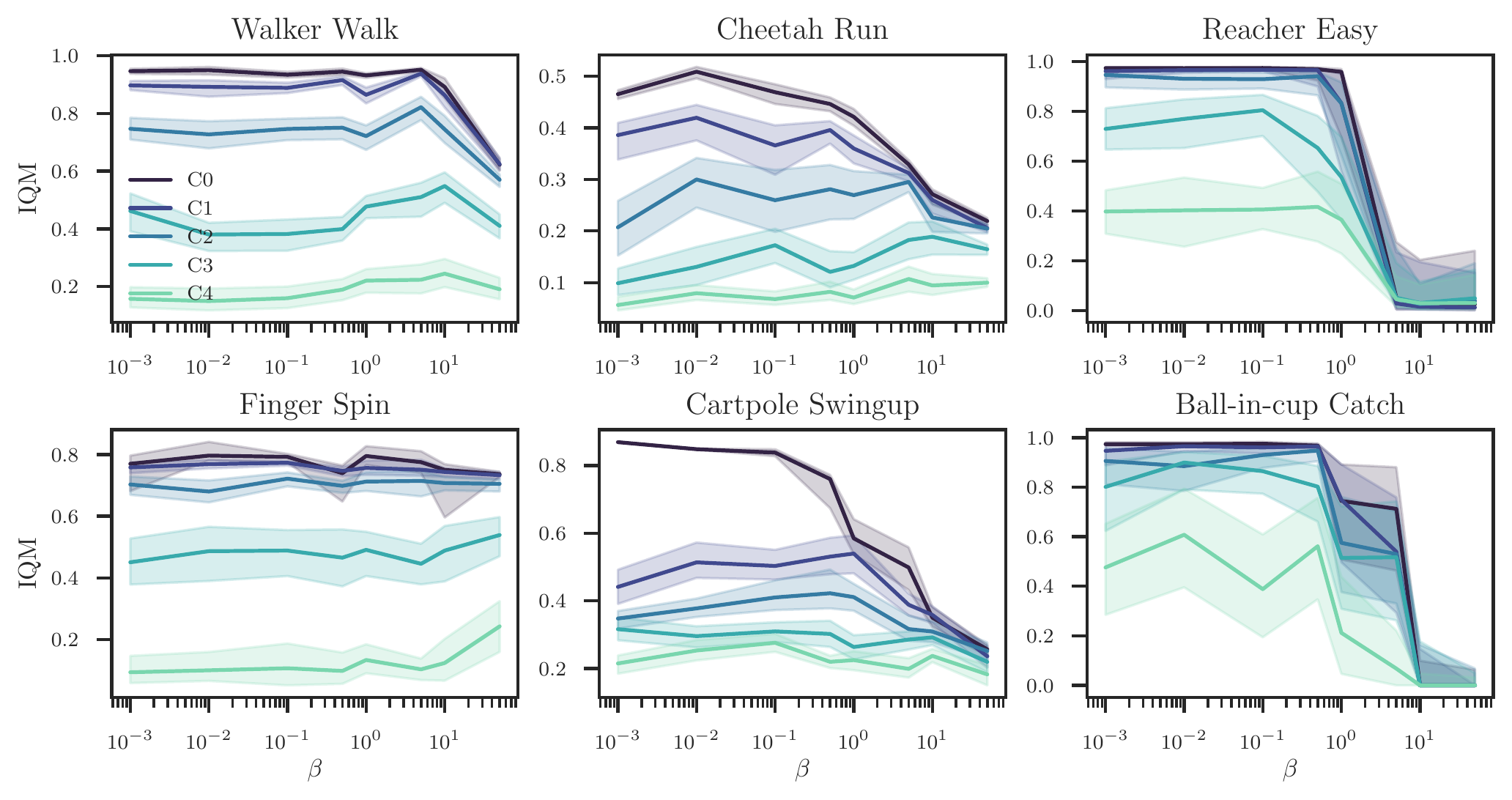}
  \end{center}
  \caption{Study of the impact of $\beta$ on generalization and invariance. Shaded areas are bootstrapped CI.}
  \label{fig:beta}
\end{figure}

\section{Continuous Control with VIBR} \label{appendix:SAC}

\paragraph{SAC}
Soft Actor-Critic \citep{haarnoja2018soft} extends Q-learning to continuous control with an entropy maximizing actor-critic algorithm. The policy loss is defined as follow:
$$
J_\pi(\phi)=\mathbb{E}_{\mathbf{s}_t \sim \mathcal{D}}\left[\mathbb{E}_{\mathbf{a}_t \sim \pi_\phi}\left[\alpha \log \left(\pi_\phi\left(\mathbf{a}_t \mid \mathbf{s}_t\right)\right)-Q_\theta\left(\mathbf{s}_t, \mathbf{a}_t\right)\right]\right]
= \mathbb{E}_{\mathbf{s}_t \sim \mathcal{D}} \left[ L_\pi(\mathbf{s}_t) \right]
$$

 As in DQN, the temporal-difference loss also minimizes Bellman residuals to learn the action-value function but with a different target: it has an additional entropy regularization term and sample the target action according to the soft-policy. 
$\mathbf{a}_{t+1} \sim \pi_\phi(\mathbf{s}_t)$:
$$
Q_{\text{target}} = r_t + \gamma \left( Q_{\bar{\theta}} (\mathbf{s}_{t+1} , \mathbf{a}_{t+1}) - \alpha \log \pi_\phi\left(\mathbf{a}_{t+1} \mid \mathbf{s}_{+1} \right) \right)
$$
where $\alpha$ is a temperature parameter controlling exploration and is either fixed or trainable.\

\paragraph{DrQ} \citep{yarats2020image} is an extension of SAC that largely improves visual RL performance on continuous control tasks. It achieves such results with random shift data augmentation, averaging the Q-target over $K$ image transformations and averaging the Q-function itself over $M$ image transformations.

\section{Representation Learning for RL}
\label{appendix:representation}

We use VIBR on top of Soft-Actor Critic \citep{haarnoja2018soft} for continuous control in the DCS environment. Our implementation follows DrQ \citep{yarats2020image} and we compare ourselves with 4 other baselines learning 
view-invariant representations:
  
 \paragraph{DBC} \citep{zhang2021learning} learns invariant representations using bisimulation metrics. It learns to put 2 distincts embedding of states at a fixed pre-computed pseudo-distance depending on behavioral similarity and optimizes the following loss function for the encoder $\phi$:
 \begin{equation}
    J(\phi)=\left(\left\|\mathbf{z}_i-\mathbf{z}_j\right\|_1-\left|r_i-r_j\right|-\gamma W_2\left(\hat{\mathcal{P}}\left(\cdot \mid \overline{\mathbf{z}}_i, \mathbf{a}_i\right), \hat{\mathcal{P}}\left(\cdot \mid \overline{\mathbf{z}}_j, \mathbf{a}_j\right)\right)\right)^2
\end{equation}
where $\mathbf{z}_i=\phi\left(\mathbf{s}_i\right), \mathbf{z}_j=\phi\left(\mathbf{s}_j\right), r$ are rewards, $\overline{\mathbf{z}}$ denotes $\phi(\mathbf{s})$ with stop gradients, $W_2$ denotes the earth-mover or 2-Wasserstein distance (which has a closed form for Gaussian distributions), and $\hat{\mathcal{P}}$ is a dynamics model with Gaussian distribution output.\

 \paragraph{SPR} \citep{schwarzer2021data} learn time predictive representations of images by predicting multiple latent vectors into the future with a siamese architecture inspired from \cite{grill2020bootstrap}. This method has a projection network and a cosine similarity loss.
 \begin{equation}
\mathcal{L}_\theta^{\mathrm{SPR}}\left(s_{t: t+K}, a_{t: t+K}\right)=-\sum_{k=1}^K\left(\frac{\tilde{y}_{t+k}}{\left\|\tilde{y}_{t+k}\right\|_2}\right)^{\top}\left(\frac{\hat{y}_{t+k}}{\left\|\hat{y}_{t+k}\right\|_2}\right)
\end{equation}

 \paragraph{CURL} is a natural extension of \cite{chen2020improved} to RL where the teacher-student architecture (with a projection network) match two different views of the same observation (originally with data augmentation) with a InfoNCE loss of the form:
 \begin{equation}
\mathcal{L}_\mathrm{CURL}=\log \frac{\exp \left(q^T W k_{+}\right)}{\exp \left(q^T W k_{+}\right)+\sum_{i=0}^{K-1} \exp \left(q^T W k_i\right)}
\end{equation}

 \paragraph{FM(Feature-Matching)} is a very simple representation learning baselines adapted to our case, where we directly match representations of two observations from the same state given by two observers $x^k$ and $x^l$. The loss (MSE) directly optimizes the encoder without projection network.
\begin{equation}
    \mathcal{L}_\mathrm{FM} = \lVert \phi(x^k(s)) - \phi(x^l(s))\rVert^2
\end{equation}

For fairness of comparison, all baselines have access to the same training data as VIBR, and might freely benefit from having multiple observers the same way as VIBR does. We maximize their performance by matching two views from different observers in each baseline's respective loss using the two-branch teacher-student architecture.

\section{Loss Landscape Toy Experiment}
\label{appendix:toyExp}
We create the plots of Figure \ref{fig:loss_landscape} by creating a pseudo loss landscape of a 2d neural network with the following bivariate function:
\begin{equation}
    f(x,y,a,b,c,d)= \left( \frac{x-a}{c} \right)^2 + \left( \frac{y-b}{c} \right)^2 + d
\end{equation}
We implement the four domains by plotting 4 variation of $f$ along the range of parameters with the following values for $a,b,c$ and $d$:
\begin{itemize}
    \item Bottom-left $\mathcal{D}^1$: $a=b=c=1 \qquad d=0$
    \item Bottom-right $\mathcal{D}^2$: $a=2 \qquad b=1 \qquad c=0.75 \qquad d=0.25$
    \item Top-left $\mathcal{D}^3$: $a=1 \qquad b=2 \qquad c=0.75 \qquad d=0.25$
    \item Top-right $\mathcal{D}^4$: $a=2 \qquad b=2 \qquad c=0.5 \qquad d=0.4$
\end{itemize}
This parameters allows us to control the width and depth of each valley, which helps us simulate different training landscapes with different optimization difficulties. We apply $\tanh$ before plotting for better visualization. The plot on the left is obtained by taking $\min_{\mathcal{D}^i} f(x,y,\mathcal{D}^i)$ for all $x,y$. This is a practical way of visualizing all 4 domains at the same time, but does not reflect a single optimization objective. The middle plot is obtained with: $\widehat{\E}_{\mathcal{D}^i} \left[ f(x,y,\mathcal{D}^i) \right]$ for all $x,y$. This corresponds to ERM if we suppose that the cardinality of each domain is equal. Finally, we obtain the plot on the right with $\widehat{\E}_{\mathcal{D}^i} \left[ f(x,y,\mathcal{D}^i) \right] + \beta \widehat{Var}\left( f(x,y,\mathcal{D}^i) \right)$ for all $x,y$. This effectively corresponds to V-Rex in \cite{krueger2021out} and our variance regularization term of VIBR.

\section{Implementation Details} \label{appendix:implementation}

Each experience in the paper is run on 4 different seeds for reproducibility. 
We base our VIBR implementation of the SAC implementation in  ACME\footnote{\url{https://github.com/deepmind/acme}} \citep{hoffman2020acme} 
in Jax \citep{jax2018github}, but modify it to fit DrQ architecture. 
Both policy and Q-network are implemented with convolutional stack followed by a MLP. 
The policy and Q-network only share weights of the convolution stack to compute lower-dimensional 
visual features. The convolutional stack (or "encoder) is composed of 4 convolutional 
layers with 32 filters and $3 \times 3$ kernel sizes. Stride is 2 for the first convolutional 
layer then 1 for the rest. Outputs features are flattened and put through a linear 
layer to reach a final dimension of 50. Layer normalization and tanh activation is applied 
to the features before passing them to actor or critic's MLP. Encoder layers are 
initialized with delta orthogonal initialization, while all linear layers used Lecun 
uniform initialization. All networks use ReLU activations units. Trainig is done with 
the Adam optimizer, and all hyperparameters used are described in table \ref{tables:SAChp}. 
We optimize $\beta$ with hyperparameter search over $\left[10^{-3},\cdots,50 \right]$ and find a value that maximizes IQM and transfer for each environment. Values are listed in the hyperparameter table. \ 

We use the github\footnote{\url{https://github.com/geyang/gym-distracting-control}} 
implementation of the Distracting Control suite and modify it to create our training and evaluation curriculum.

\begin{table}[!htb]
  \caption{Hyperparameters}
  \label{tables:SAChp}
  \centering
  \begin{tabular}{lc}
    \toprule
    Hyperparameter     & Value \\
    \midrule
    Replay buffer size & 100000 \\
    Initial collection steps & 25000 \\
    Optimizer & Adam \\
    Actor learning rate & 3e-4 \\
    Critic learning rate & 3e-4 \\
    Weight decay & 0 \\
    Initial temperature $\alpha$ & 0.1 \\
    Temperature learning rate & 3e-4 \\
    Batch size & 128 \\
    $\tau$ EMA & 5e-3 \\
    Actor hidden layers & [512, 512] \\
    Critic hidden layers & [512, 512] \\
    Frame stacking & 3 \\
    Action repeat & 8 if Cartpole; 2 if {Finger, Walker}; 4 otherwise \\
    VIBR variance penalty $\beta$ & 5 if Walker; 0.1 if Ball in Cup; 1 otherwise \\
    \bottomrule
  \end{tabular}
\end{table}

\FloatBarrier
\section{Distracting Control Suite}
\label{appendix:DCS}
\begin{figure}[h!]
    \label{fig:DCS}
  \begin{center}
    \includegraphics[width=\textwidth]{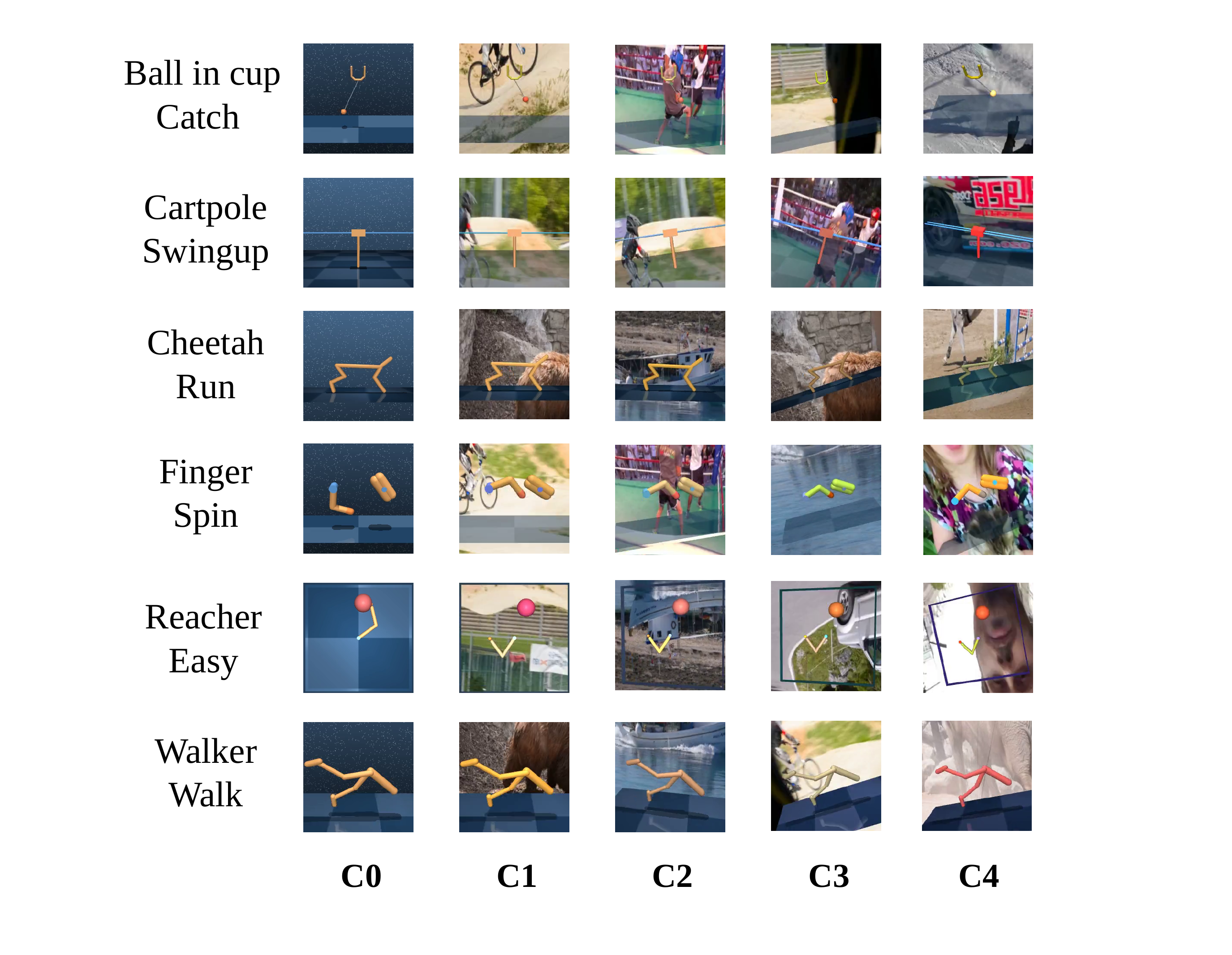}
  \end{center}
  \caption{Distracting Control Suite tasks and evaluation benchmarks used in the paper.}
\end{figure}

We use Distracting Control Suite \citep{stone2021distracting} for our experiments. 
DCS is a variant of the Deepmind Control Suite where visual distractions are dynamically added to the rendered observations.
The perturbations consists in the following non-exclusive dimension of variations: 
\begin{itemize}
    \item color randomization of physical bodies
    \item background randomization with a dataset of videos
    \item random camera wobbling around a fixed point
\end{itemize}
Distractions are dynamic, temporally consistent and continuous. 
Colors of bodies are continuously changing at each time step. The background is displaying frame by frame a randomly selected 
video from the DAVIS dataset, which is played forward then backward to avoid discontinuities. The camera's orientation is 
rotating with a random angle at each step while keeping the agent in the field of view. We define our curriculum as follows:
\begin{itemize}
    \item \textbf{C0}: No visual perturbation, original DM Control environment
    \item \textbf{C1}: Dynamic background changes with a dataset of 2 videos and random body colorization with an intensity parameter $\alpha$ of 0.1
    \item \textbf{C2}:  Dynamic background changes with a dataset of 4 videos, random body colorization and random camera wobbling with an intensity parameter $\alpha$ of 0.1
    \item \textbf{C3}: Dynamic background changes with a dataset of 8 videos, random body colorization and random camera wobbling with an intensity parameter $\alpha$ of 0.2
    \item \textbf{C4}:  Dynamic background changes with a dataset of 50 videos, random body colorization and random camera wobbling with an intensity parameter $\alpha$ of 0.3
\end{itemize}
Training and evaluation curriculums have the same definition, except that training domains use a separate video dataset than evaluation domains to properly evaluate generalization both in-distribution and out-of-distribution.
All models are consistently trained with both C0 and either C1, C2 or C3. Main results are obtained with C0 and C2. This allows us to easily define 2 observers $x^0$ and $x^2$ where $x^0(\mathcal{S}) \in C0 $ and $x^2(\mathcal{S}) \in C2 $.
Using these two observers, we produce 2 different view each on its own domain at each time step. We provide image samples of each environment for each domain of the curriculum in Figure \ref{fig:DCS}

\section{Sample efficiency curves} \label{appendix:efficiency}

\begin{figure}[h!]
  \begin{center}
    \includegraphics[width=\textwidth]{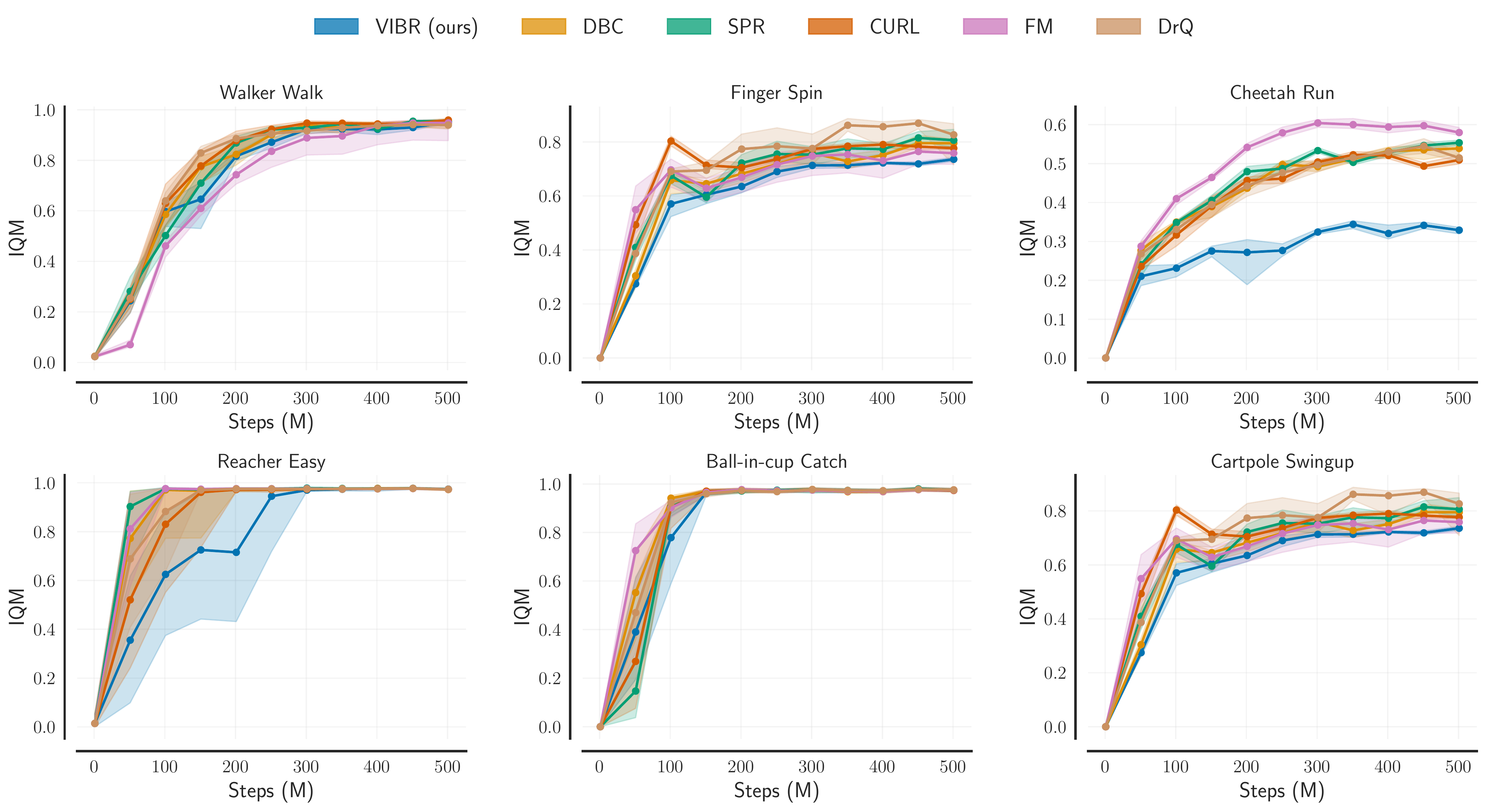}
  \end{center}
  \caption{Evaluation sample efficiency curves on C0. Line IQM and shaded area bootstrapped CI.}
\end{figure}

\begin{figure}[h!]
  \begin{center}
    \includegraphics[width=\textwidth]{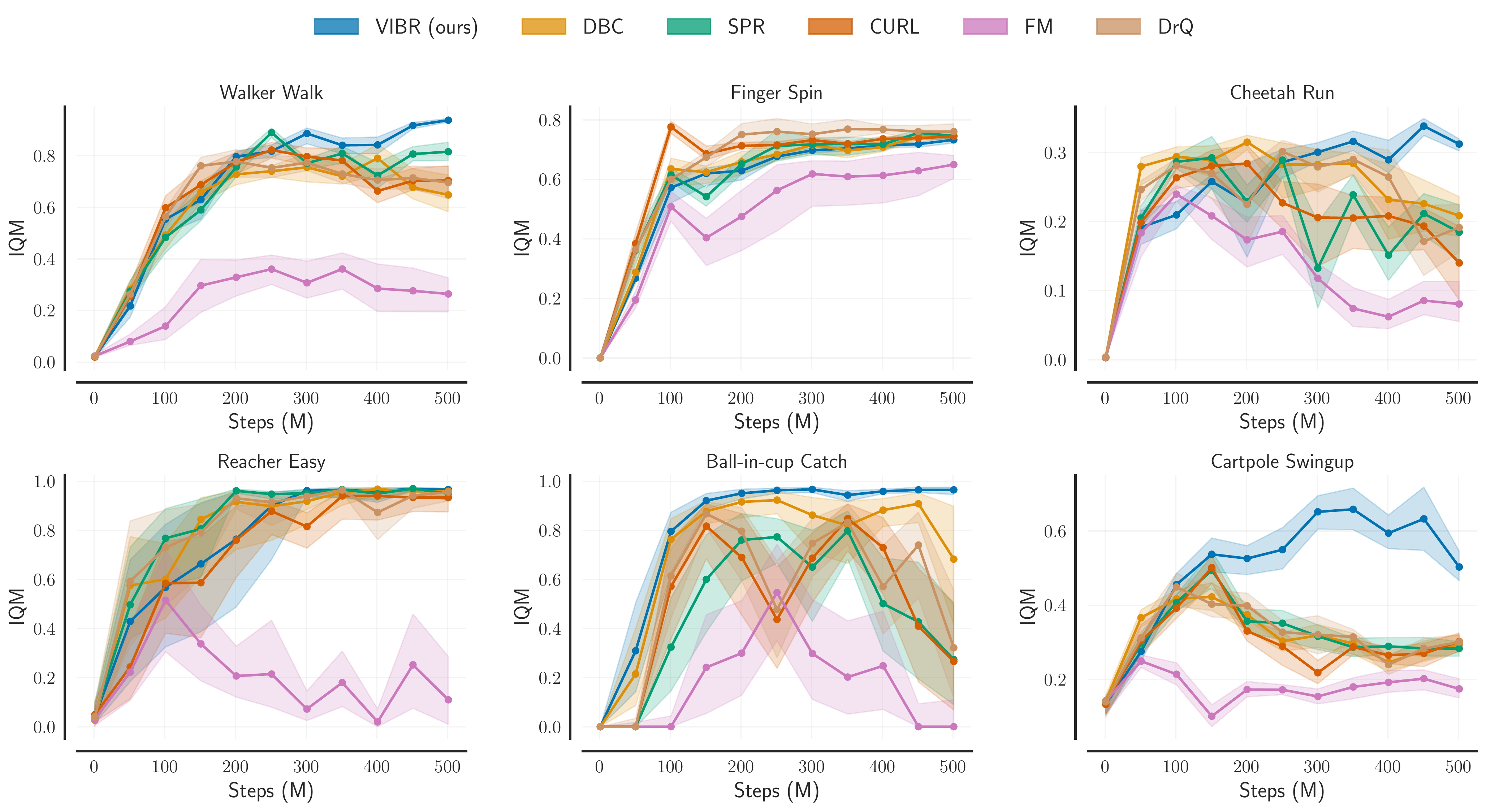}
  \end{center}
  \caption{Evaluation sample efficiency curves on C1. Line IQM and shaded area bootstrapped CI.}
\end{figure}

\begin{figure}[h!]
  \begin{center}
    \includegraphics[width=\textwidth]{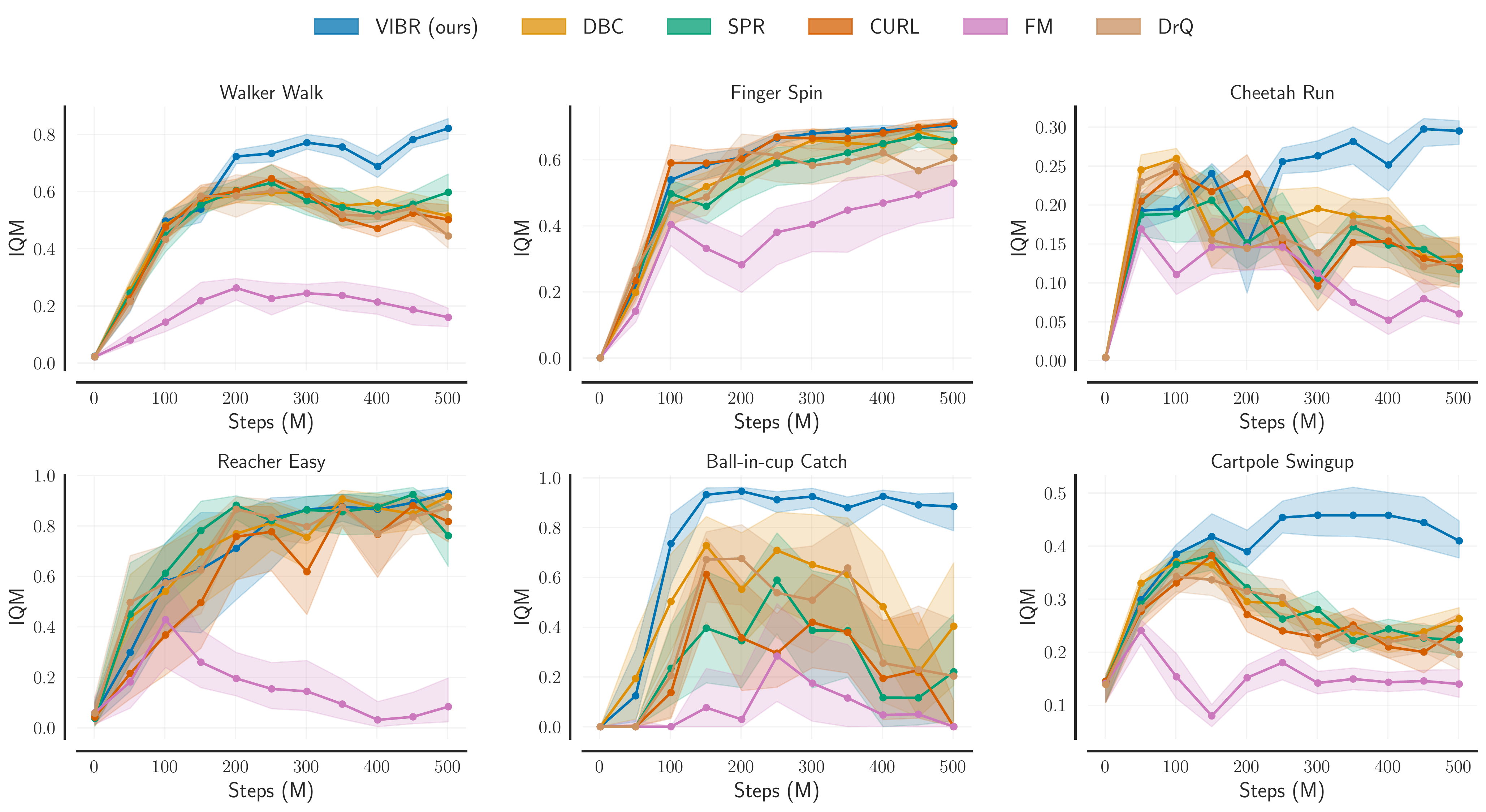}
  \end{center}
  \caption{Evaluation sample efficiency curves on C2. Line IQM and shaded area bootstrapped CI.}
\end{figure}

\begin{figure}[h!]
  \begin{center}
    \includegraphics[width=\textwidth]{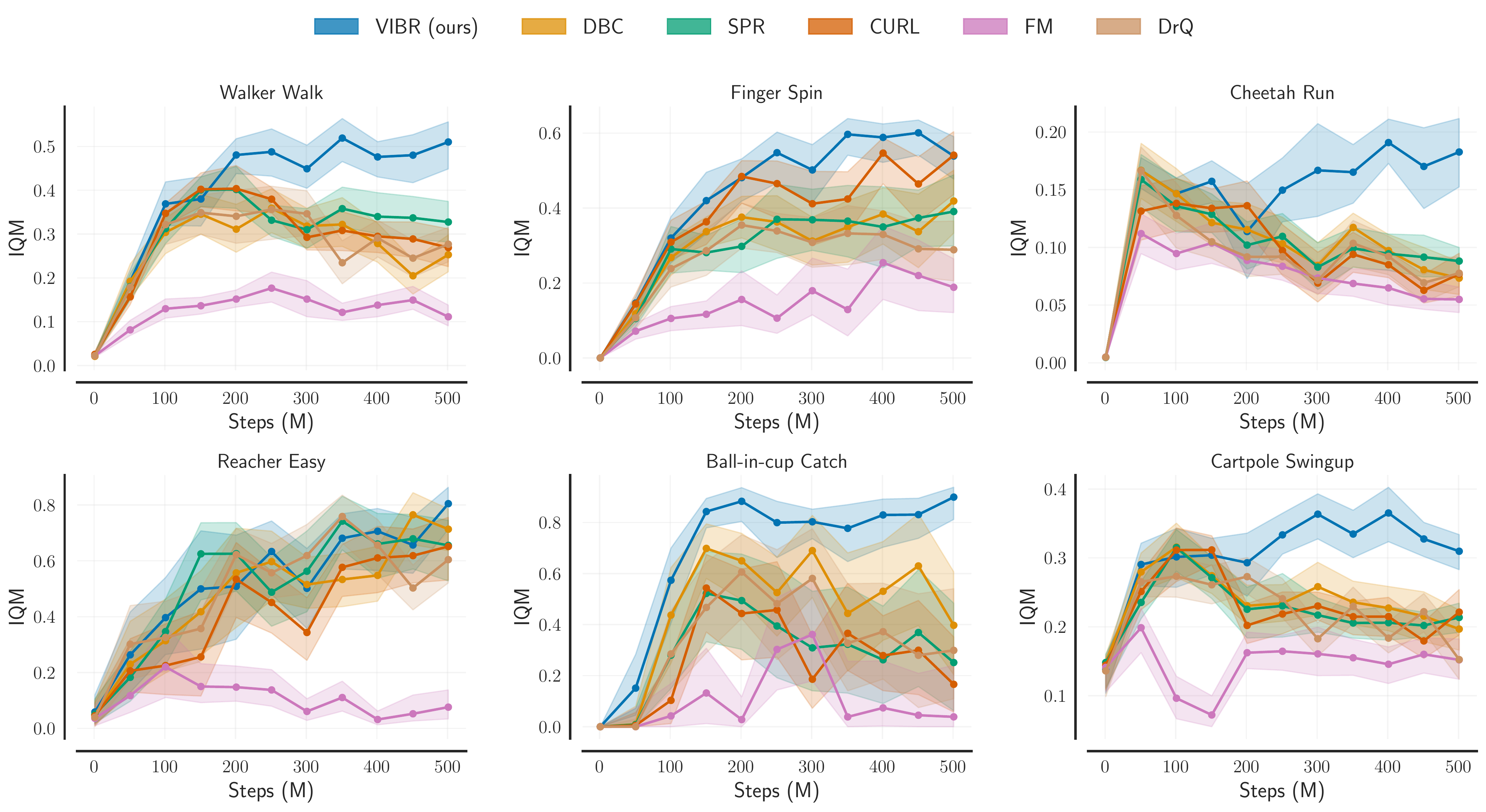}
  \end{center}
  \caption{Evaluation sample efficiency curves on C3. Line IQM and shaded area bootstrapped CI.}
\end{figure}

\begin{figure}[h]
  \begin{center}
    \includegraphics[width=\textwidth]{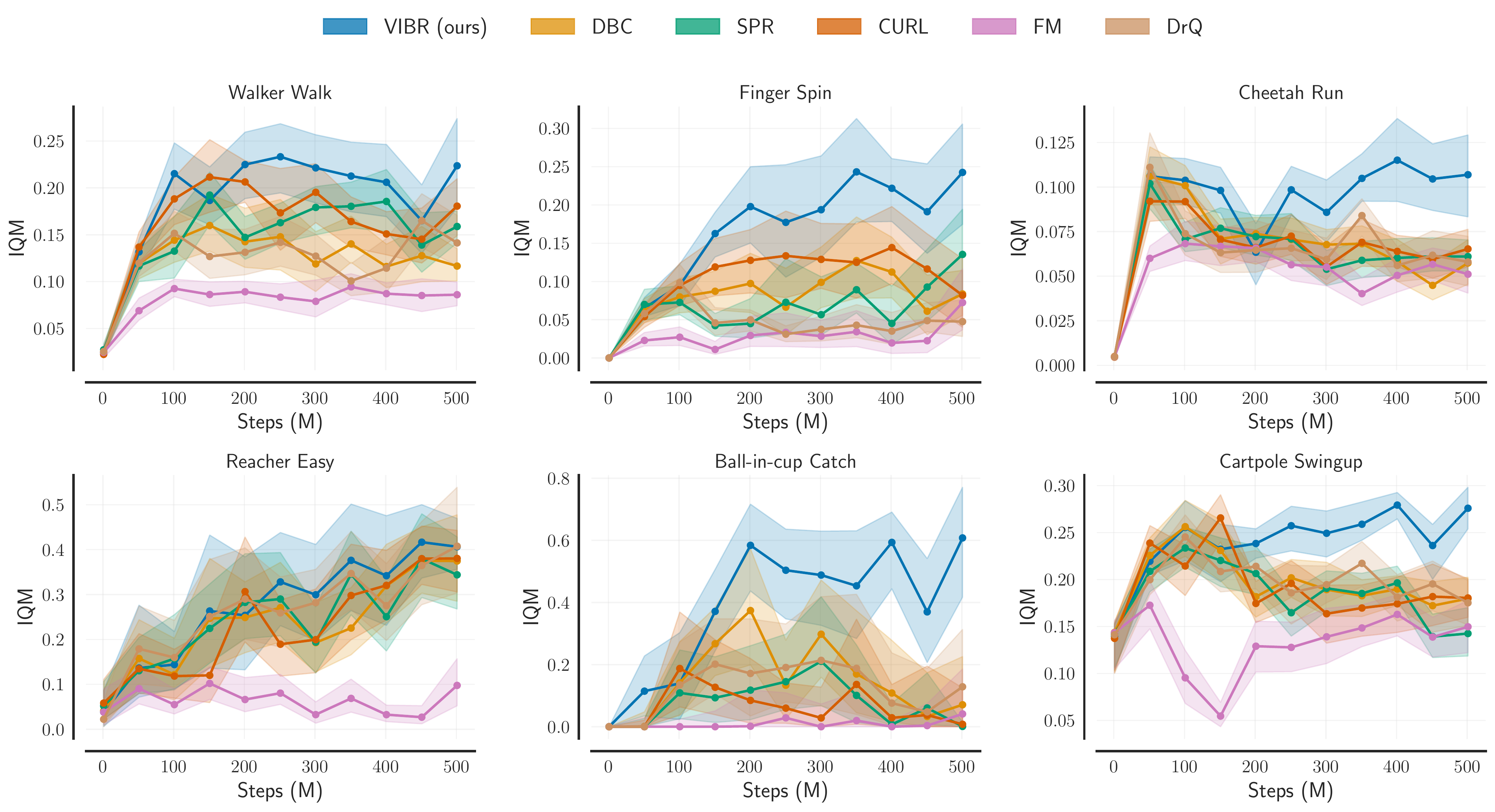}
  \end{center}
  \caption{Evaluation sample efficiency curves on C4. Line IQM and shaded area bootstrapped CI.}
\end{figure}

\FloatBarrier

\section{IQM and Generalization Gap per Environment per Benchmark}  \label{appendix:iqm}

\begin{figure}[h!]
  \begin{center}
    \includegraphics[width=\textwidth]{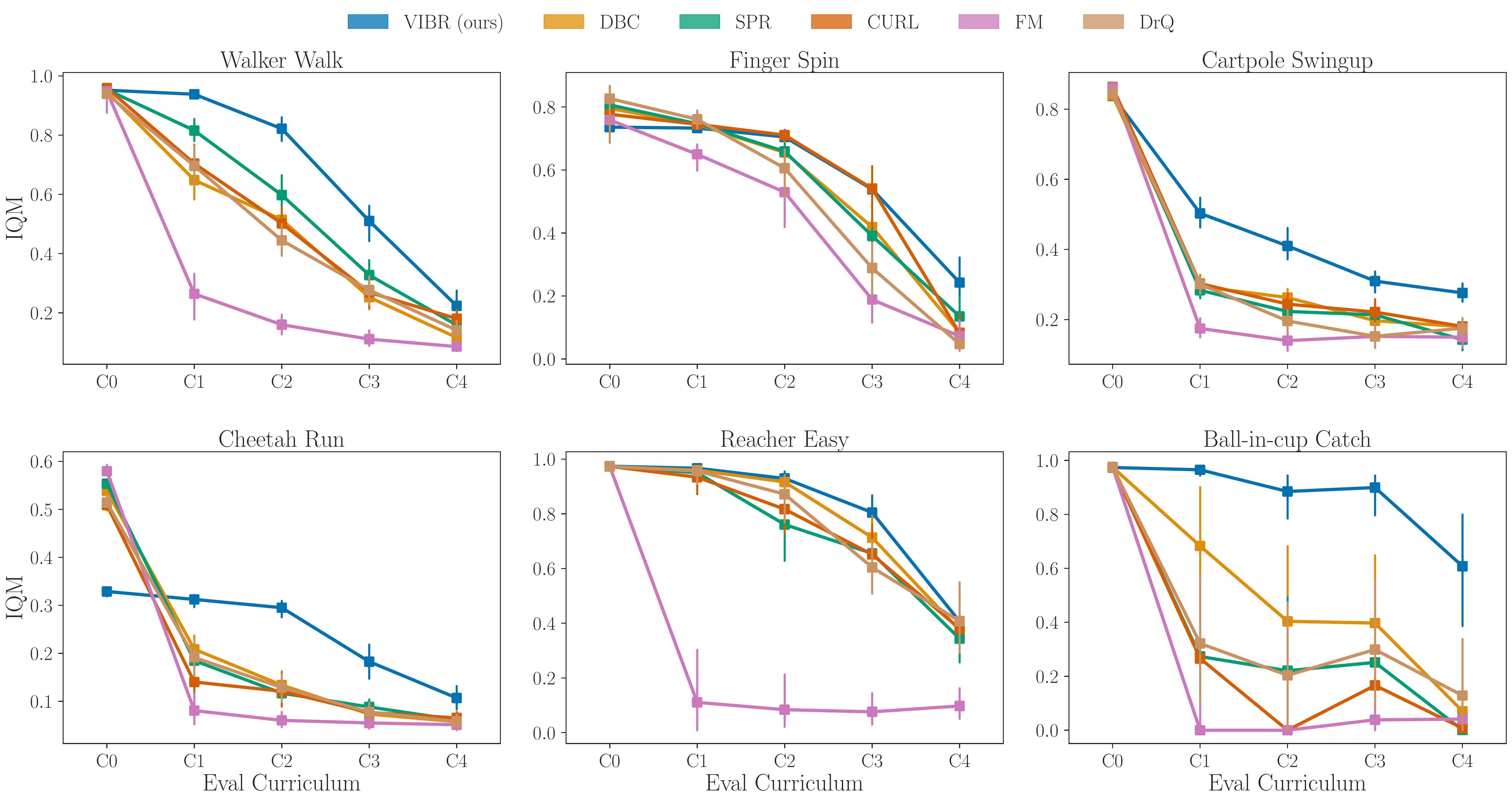}
  \end{center}
  \caption{IQM per environment per evaluation domain compared to baselines. Trained on C2*. Vertical bars are bootstrapped CI.}
\end{figure}

\begin{figure}[h!]
  \begin{center}
    \includegraphics[width=\textwidth]{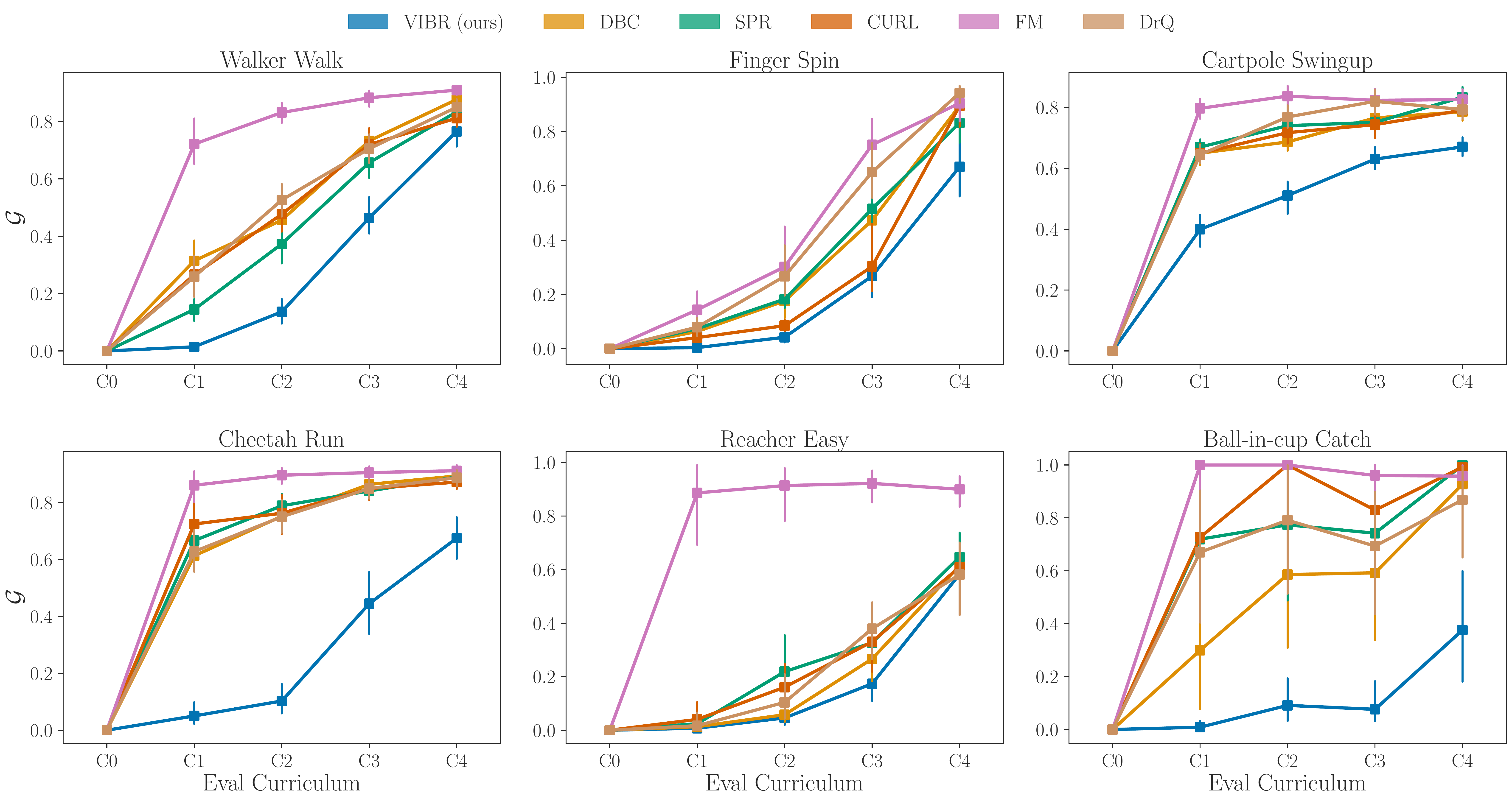}
  \end{center}
  \caption{Generalization Gap per environment per evaluation domain compared to baselines. Trained on C2*. Vertical bars are bootstrapped CI.}
\end{figure}

\begin{figure}[h!]
  \begin{center}
    \includegraphics[width=\textwidth]{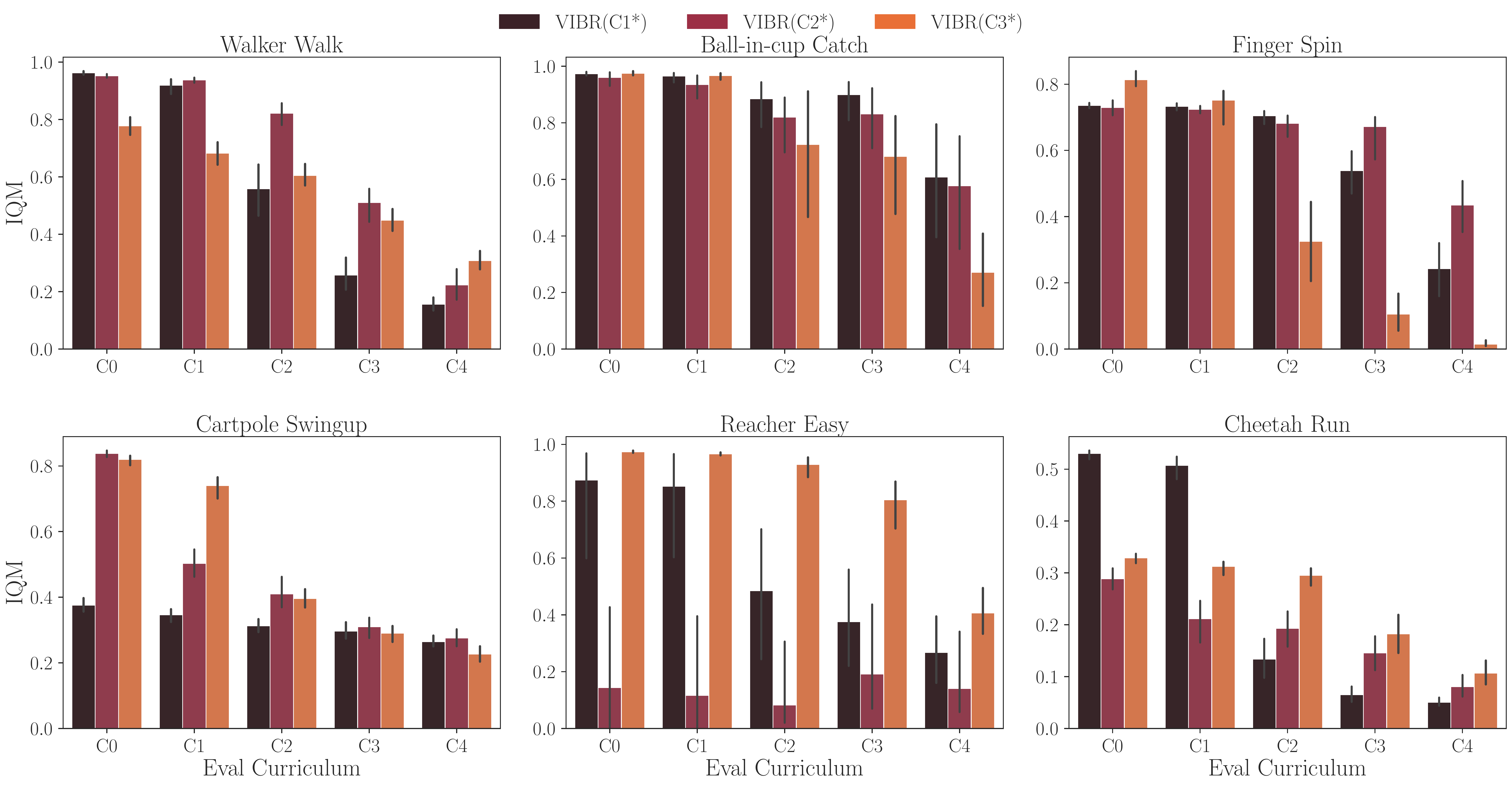}
  \end{center}
  \caption{IQM per environment per evaluation domain while changing training domain. Vertical bars are bootstrapped CI.}
\end{figure}

\begin{figure}[h!]
  \begin{center}
    \includegraphics[width=\textwidth]{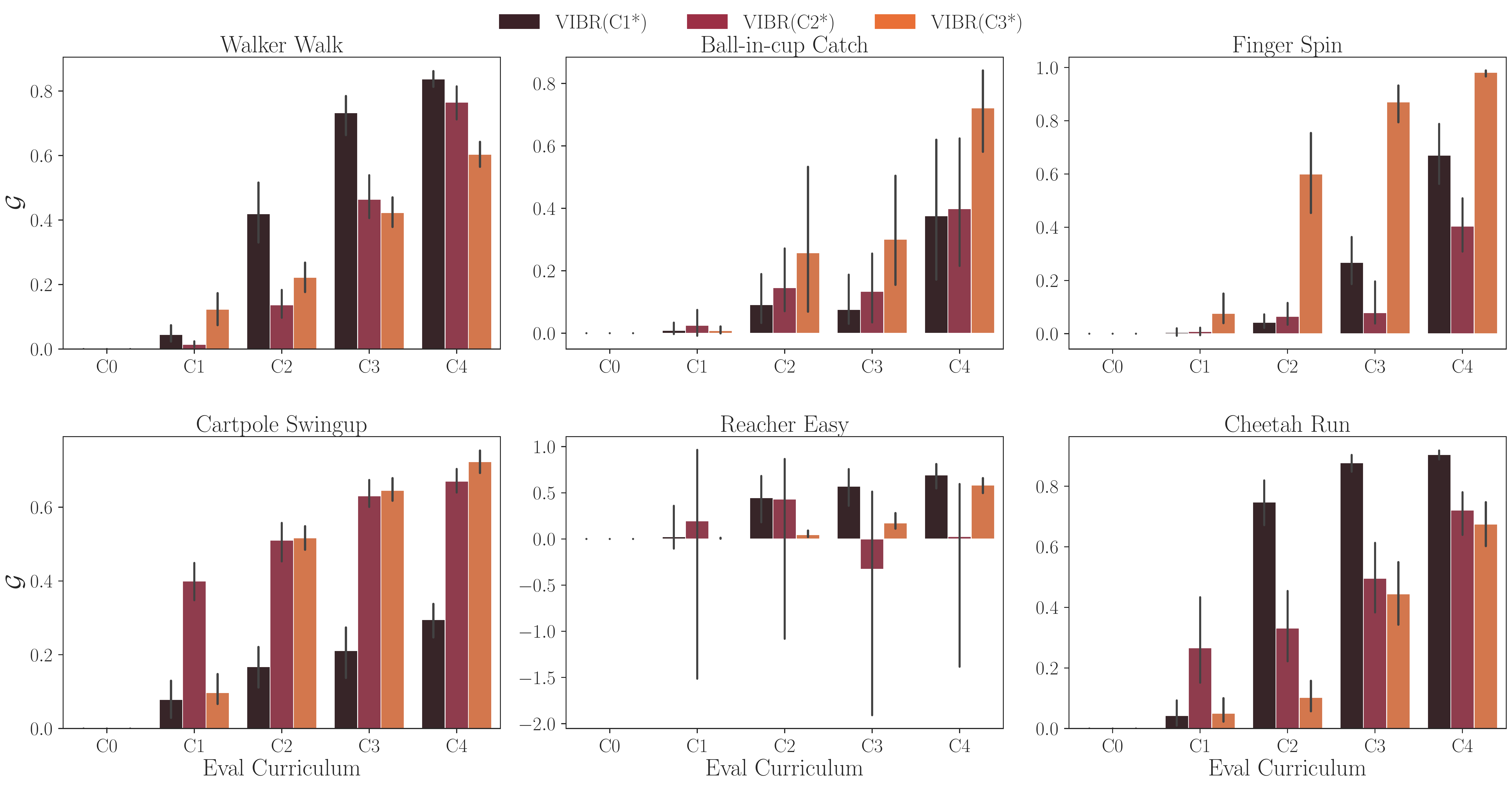}
  \end{center}
  \caption{Generalization Gap per environment per evaluation domain while changing training domain. Vertical bars are bootstrapped CI.}
\end{figure}

\FloatBarrier

\section{Ablations} \label{appendix:ablation}
We perform ablations on the $\widehat{\E}\left[ \mathcal{L}_\mathrm{BR}(k,l)\right]$ term from the VIBR loss and show that all Bellman residuals terms are necessary for good performance. To perform the ablation, we disabled risk extrapolation by fixing $\beta$ to 0.  Training is done on the C2* on the Walker Walk environment for 4 seeds. 

\begin{figure}[h!]
  \begin{center}
    \includegraphics[width=.7\textwidth]{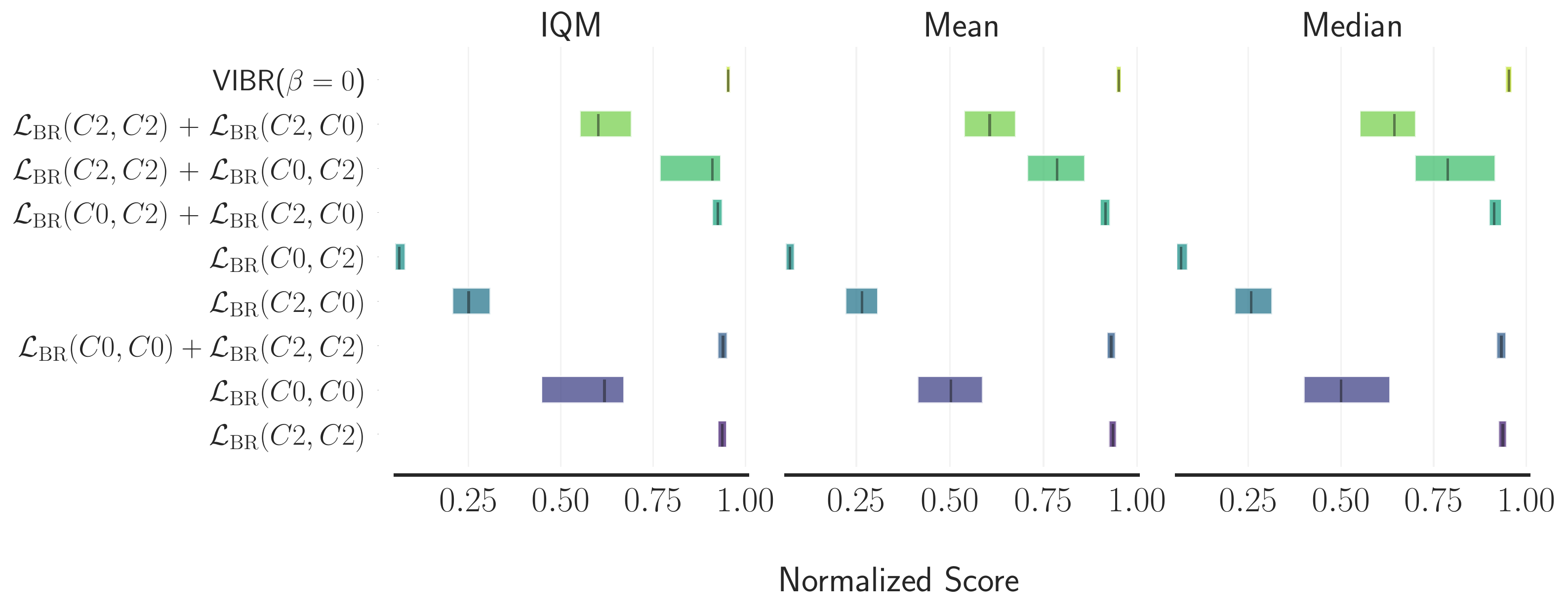}
  \end{center}
  \caption{IQM and bootstrapped CI of ablations on C0}
\end{figure}

\begin{figure}[h!]
  \begin{center}
    \includegraphics[width=.7\textwidth]{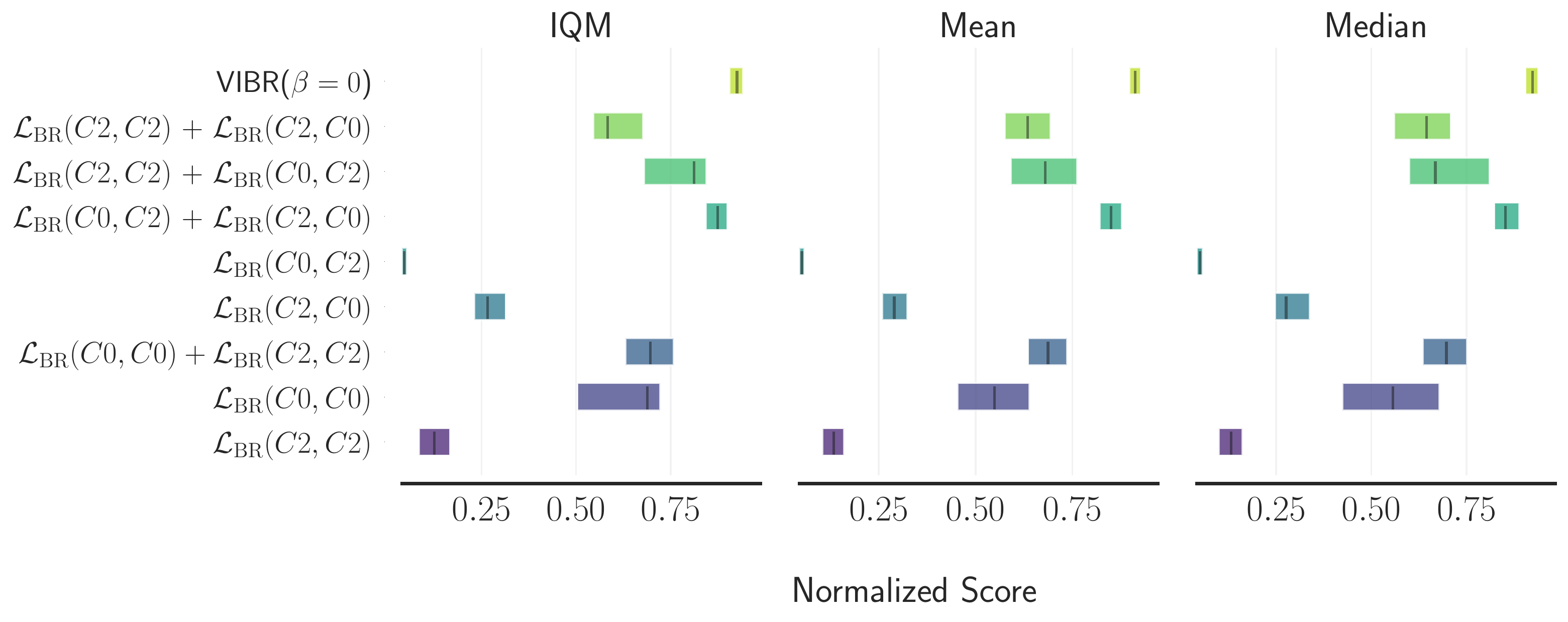}
  \end{center}
  \caption{IQM and bootstrapped CI of ablations on C1}
\end{figure}

\begin{figure}[h!]
  \begin{center}
    \includegraphics[width=.7\textwidth]{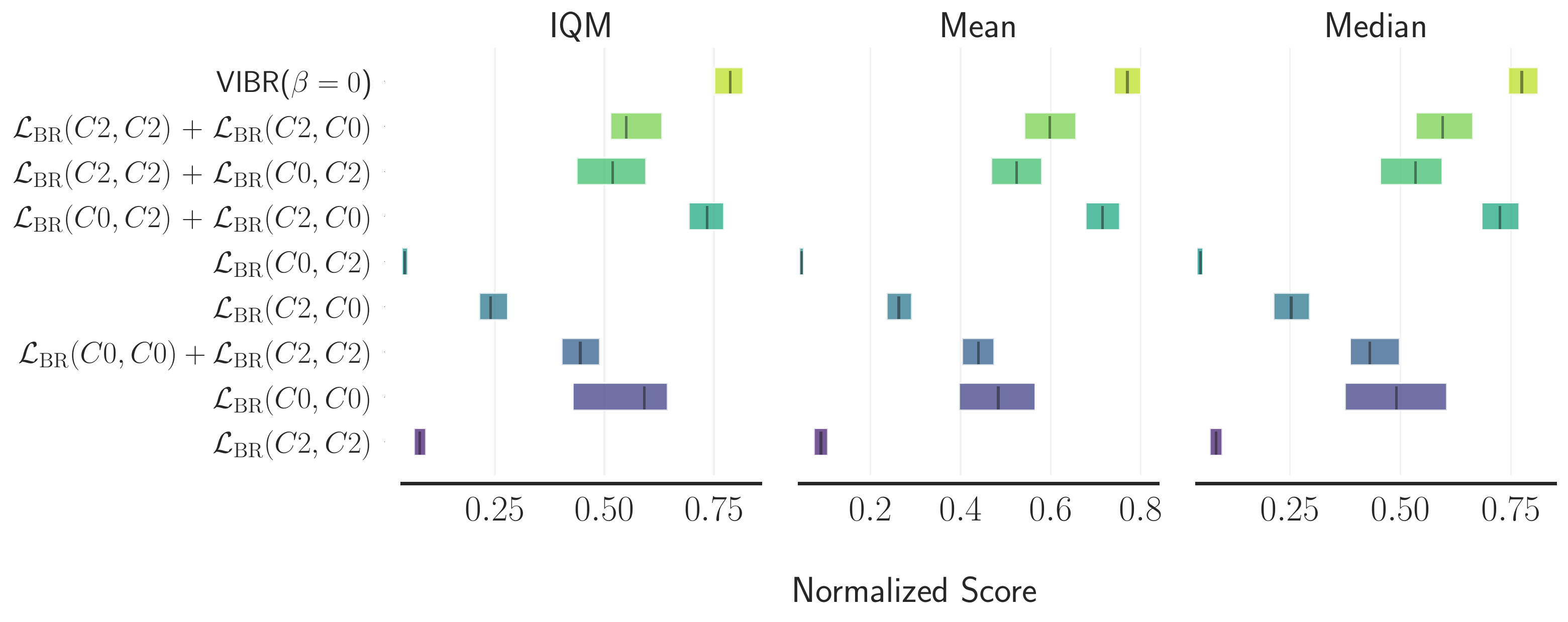}
  \end{center}
  \caption{IQM and bootstrapped CI of ablations on C2}
\end{figure}

\begin{figure}[h!]
  \begin{center}
    \includegraphics[width=.7\textwidth]{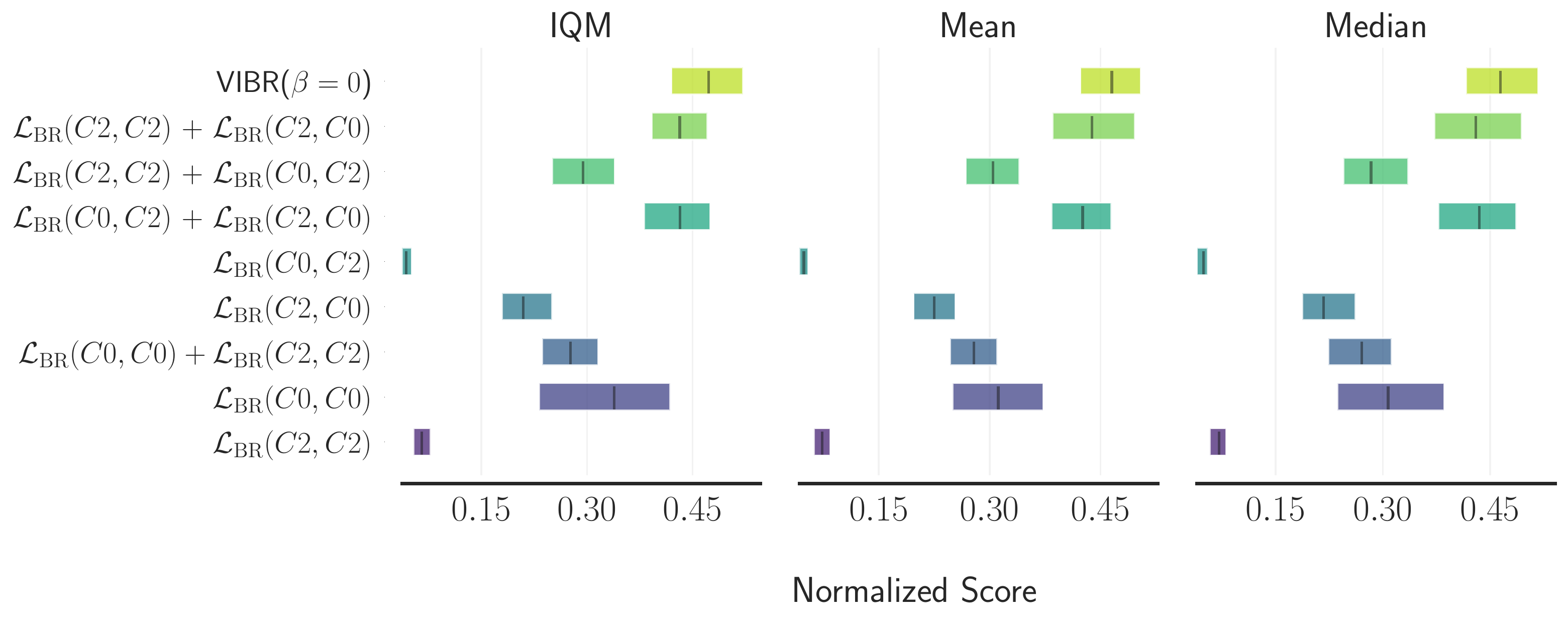}
  \end{center}
  \caption{IQM and bootstrapped CI of ablations on C3}
\end{figure}

\begin{figure}[h!]
  \begin{center}
    \includegraphics[width=.7\textwidth]{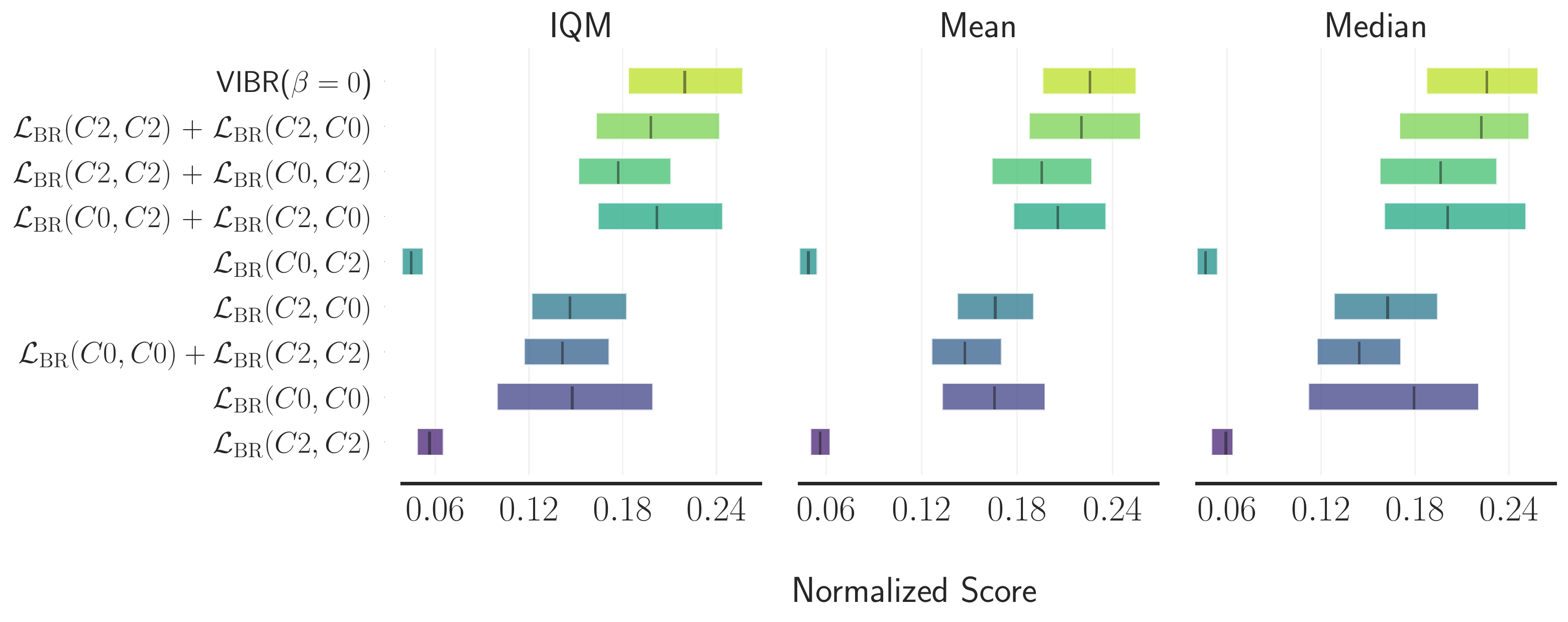}
  \end{center}
  \caption{IQM and bootstrapped CI of ablations on C4}
\end{figure}

\end{document}